\documentclass[10pt,twocolumn,letterpaper]{article}
\usepackage[pagenumbers]{cvpr} 

\usepackage[utf8]{inputenc} 
\usepackage{url}            
\usepackage{booktabs}       
\usepackage{amsfonts}       
\usepackage{amsmath}       %
\usepackage{nicefrac}       
\usepackage{microtype}      
\usepackage{xcolor}         
\usepackage{gensymb}
\usepackage{bm}
\usepackage{amssymb}
\usepackage{subcaption}
\usepackage{graphicx}
\usepackage{multirow}

\definecolor{cvprblue}{rgb}{0.21,0.49,0.74}
\usepackage[pagebackref,breaklinks,colorlinks,allcolors=cvprblue]{hyperref}

\title{OmniSplat: Taming Feed-Forward 3D Gaussian Splatting \\ for Omnidirectional Images with Editable Capabilities}

\author{
  Suyoung Lee\thanks{indicates equal contribution.}~~$^1$ \qquad Jaeyoung Chung\footnotemark[1]~~$^1$ \qquad Kihoon Kim~$^2$ \qquad Jaeyoo Huh~$^2$ \\
  Gunhee Lee~$^3$ \qquad Minsoo Lee~$^3$ \qquad Kyoung Mu Lee~$^{1,2}$ \\
  $^1$Dept. of ECE \& ASRI, $^2$IPAI, Seoul National University, $^3$LG AI Research, Seoul, Korea\\
  \texttt{\{esw0116,\,robot0321,\,kihoon96\}@snu.ac.kr}\qquad \texttt{jaeyoo900@gmail.com} \\ \texttt{\{gunhee.lee,\,minsoo.lee\}@lgresearch.ai}\qquad\texttt{kyoungmu@snu.ac.kr} \\
}

\begin{document}
\maketitle
\begin{abstract}
Feed-forward 3D Gaussian splatting (3DGS) models have gained significant popularity due to their ability to generate scenes immediately without needing per-scene optimization.
Although omnidirectional images are becoming more popular since they reduce the computation required for image stitching to composite a holistic scene, existing feed-forward models are only designed for perspective images.
The unique optical properties of omnidirectional images make it difficult for feature encoders to correctly understand the context of the image and make the Gaussian non-uniform in space, which hinders the image quality synthesized from novel views.
We propose OmniSplat, a training-free fast feed-forward 3DGS generation framework for omnidirectional images.
We adopt a Yin-Yang grid and decompose images based on it to reduce the domain gap between omnidirectional and perspective images.
The Yin-Yang grid can use the existing CNN structure as it is, but its quasi-uniform characteristic allows the decomposed image to be similar to a perspective image, so it can exploit the strong prior knowledge of the learned feed-forward network.
OmniSplat demonstrates higher reconstruction accuracy than existing feed-forward networks trained on perspective images.
Our project page is available on: \url{https://robot0321.github.io/omnisplat/index.html}.
\end{abstract}
    
\section{Introduction}
\label{sec:intro}

\begin{figure}[t]
   \centering
   \includegraphics[width=1\linewidth]{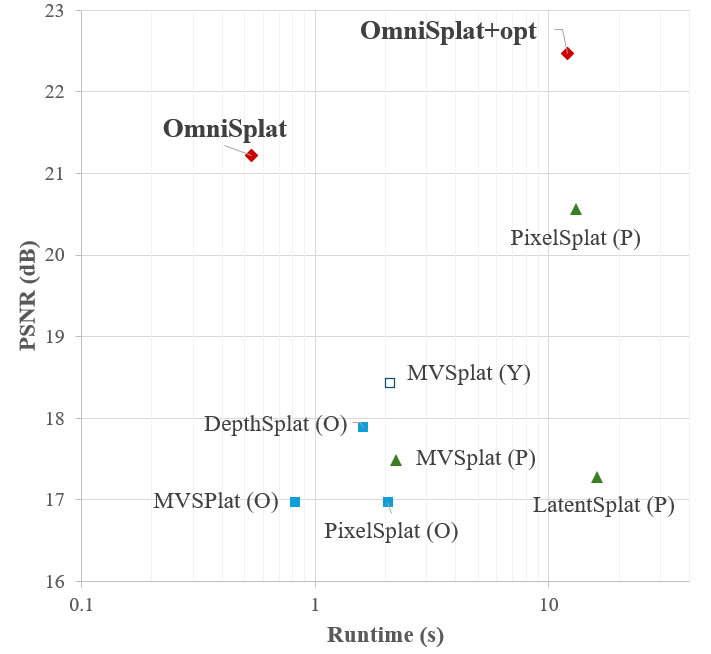}
   \caption{
        \textbf{PSNR-runtime trade-off.} A chart of reconstruction PSNR-runtime trade-off in novel view image on OmniBlender~\cite{choi2023balanced}.
        OmniSplat shows the best trade-off compared to the original feed-forward networks for perspective images.
   }
   \label{fig:teaser}
\end{figure}

The reconstruction of holistic 3D scenes from multiview images is one of the fundamental problems in computer vision with emerging applications such as virtual reality (VR), augmented reality (AR), robotics, or autonomous navigation. The goal is to rapidly and accurately create holistic 3D representations of environments. Recent advancements have focused on feed-forward scene generation networks~\cite{charatan2024pixelsplat, chen2024mvsplat, wewer24latentsplat}, which are capable of generating 3D Gaussian splatting (3DGS) representations directly from a few input images without scene-wise optimization~\cite{kerbl20233d}. These models estimate plausible 3D Gaussian parameters by leveraging priors learned from large-scale datasets and achieve more than 30 times faster than optimization-based methods.
However, they often encounter challenges in constructing a holistic scene due to the perspective camera with a limited field of view.
Constructing a holistic scene using multiple perspective images in a pair-wise manner incurs significant drawbacks in terms of both computational efficiency and reconstruction quality.
Instead, omnidirectional images have become increasingly prevalent due to their ability to capture a wide field of view within a single image.
They are computationally efficient as they reduce the amount of data and computation for image stitching to represent an entire scene compared to perspective images.
Despite the advantages, no attempts are made to estimate the parameters of 3D Gaussians directly in the omnidirectional image domain due to insufficient omnidirectional multiview image data for training the Transformer-based network.

Leveraging advanced pre-trained networks offers an effective way to overcome such data insufficiency, though several challenges remain to be addressed.
Standard omnidirectional(or equirectangular) images or fisheye images cannot be directly processed with pre-trained networks, due to their non-uniform structure.
For instance, the horizontal length of the object is stretched in polar regions of omnidirectional images, making the shape of the object different from perspective images.
Thus, the existing pre-trained network trained with perspective images often miscomprehends the context in the omnidirectional images, extracting unintended features.
In addition, such non-uniform grids promote uneven Gaussian generation, which degrades the quality of novel view synthesis.
When synthesizing an omnidirectional image from a novel view using non-uniformly distributed Gaussians, the high sampling frequency near the poles results in stripe-like artifacts.

To address these challenges, we propose \textbf{OmniSplat}, the first feed-forward 3DGS estimation from a few omnidirectional images.
We decompose each omnidirectional image into two images using Yin-Yang grid~\cite{kageyama2004yin}.
The Yin-Yang decomposition is conducted by cutting a sphere into two pieces, similar to the threads of a tennis ball, and arranging them on a plane.
Compared to other spherical representation methods such as equirectangular, fish-eye, icosahedral~\cite{iso2019} or cubed-sphere~\cite{Monroy2018cube}, the Yin-Yang grid has two advantages: \textit{quasi-uniformness} and \textit{structured grid}.
First, the quasi-uniformness of the Yin-Yang grid greatly reduces the distortion of omnidirectional images caused by equirectangular projection, making the image much more similar to perspective images.
Moreover, Yin-Yang's structured grid can exploit the strong prior of the existing pre-trained feed-forward networks, whereas the other shapes, like icosahedrons, have a different topology from typical images, making it hard to utilize the power of existing pre-trained models.
After the Gaussians are created in the space, we propose a Yin-Yang rasterizer to render Yin and Yang images for a novel view.
Then, the two rasterized images are transformed and combined into the final omnidirectional image in a pixel space.
This eliminates many artifacts caused by high sampling frequency in polar regions when using an omnidirectional rasterizer~\cite{lee2024odgs}.

Extensive experiments validate the effectiveness of our method, showing that it outperforms both the optimization-based omnidirectional 3DGS method (ODGS) and typical feed-forward generation networks in novel-view omnidirectional image reconstruction.
As described in \Cref{fig:teaser}, OmniSplat shows the fastest synthesis speed while reaching the highest PSNR than any other models.
Also, with a small number of optimizations, OmniSplat+\textit{opt} shows an overwhelming performance compared to other networks.

Additionally, we observed that the proposed architecture is well-suited for fast and efficient Gaussian segmentation. We utilize the attention scores obtained during the reconstruction process, finding multiview consistent semantic segmentation. We selected the pixel-aligned Gaussians in 3D space using this matched segmentation map. This approach results in clear boundaries without additional computation and establishes a robust foundation for 3D editing.

Our contributions can be summarized as follows:

\begin{itemize}
    \item We propose a novel method for reconstructing an entire 3D scene with a single forward step from a few omnidirectional images.
    \item By employing a Yin-Yang grid to reduce distortion and ensure uniform sampling, we successfully address the challenges associated with distortions near the poles in omnidirectional images. 
    \item Our method successfully transfers existing perspective feed-forward 3DGS estimation models to omnidirectional inputs without fine-tuning, achieving higher reconstruction accuracy compared to directly applying existing methods.
\end{itemize}
\section{Related works}
\label{sec:related}
\paragraph{Sparse view scene reconstruction and synthesis.}
The advent of Neural Radiance Fields (NeRF)~\cite{mildenhall2020nerf} and 3D Gaussian Splatting (3DGS)~\cite{kerbl20233d} has greatly advanced the field of novel view synthesis and 3D reconstruction, achieving high fidelity in scene representation when dense input views are available~\cite{lu2024scaffold,yu2024gsdf,huang2024error,lin2024vastgaussian}. 
However, capturing extensive views is often impractical in real-world settings, leading to growing interest in sparse-view approaches that aim to reconstruct from only a few input images.
Several works optimize 3DGS per scene by introducing a proper regularization that prevents the 3DGS from being overfit to training views~\cite{paliwal2024coherentgs,chung2024depth,li2024dngaussian,bao2024loopsparsegs}.
However, those models spend a long time optimizing Gaussian splats for every input image set.

The introduction of pixelNeRF~\cite{yu2020pixelnerf} marked a pivotal moment in sparse view setting, showing the advantages of feed-forward networks in terms of inference speed and the ability to leverage large-scale datasets.
Consequently, various efforts have extended feed-forward paradigms to 3DGS-based methods by regressing Gaussian parameters from pixel-aligned features~\cite{charatan2024pixelsplat, wewer24latentsplat, chen2024mvsplat,xu2024depthsplat,szymanowicz2024splatter,xu2024grm,zhang2024gslrm}.
Despite these advancements, current methods predominantly focus on conventional perspective images, limiting their application in scenarios where comprehensive scene capture is essential.
In this work, we present a novel feed-forward architecture tailored for sparse omnidirectional images, enabling fast and accurate 3D scene reconstructions. Our approach addresses the challenges of distortion inherent to omnidirectional projections and demonstrates improved performance and computational efficiency over conventional methods.

\paragraph{Omnidirectional coordinate system.} 
Inspired by the success of deep learning in visual scene recognition, recent studies have been extended to omnidirectional images, focusing initially on equirectangular projection (ERP)~\cite{coors2018sphere, renata2017graphomni, chuansu2017flat} and fisheye representations. ERP provides a simple, rectangular mapping of the sphere but introduces severe distortions at the poles, creating inconsistencies in feature extraction. Fisheye projection reduces distortion at the center of the view but experiences radial stretching at the edges, limiting its utility for consistent omnidirectional representation.

To mitigate these issues, quasi-uniform coordinate systems like icosahedral~\cite{iso2019}, cubed-sphere~\cite{Monroy2018cube}, and Yin-Yang grids~\cite{kageyama2004yin} have been introduced, aiming to better balance feature distribution. The icosahedral projection divides the sphere into 20 triangular facets, achieving near-uniform distribution but complicating compatibility with CNNs due to its triangular structure. Cubed-sphere projection aligns more closely with CNNs by dividing the sphere into six square faces, which eases integration but leads to boundary discontinuities at cube edges. In contrast, the Yin-Yang grid offers an optimal solution with two overlapping square grids that cover the sphere in a quasi-uniform manner. This structure ensures smooth transitions and seamless coverage across the sphere, minimizing polar distortion and boundary discontinuities in other methods. Its square grid design also makes it directly compatible with standard CNN operations, enabling effective feature extraction without requiring complex transformations. Here, we utilize the Yin-Yang grid to capture consistent spatial relationships vital for high-quality omnidirectional scene reconstruction and editing.
\begin{figure*}[ht]
   \centering
   \includegraphics[width=1\linewidth]{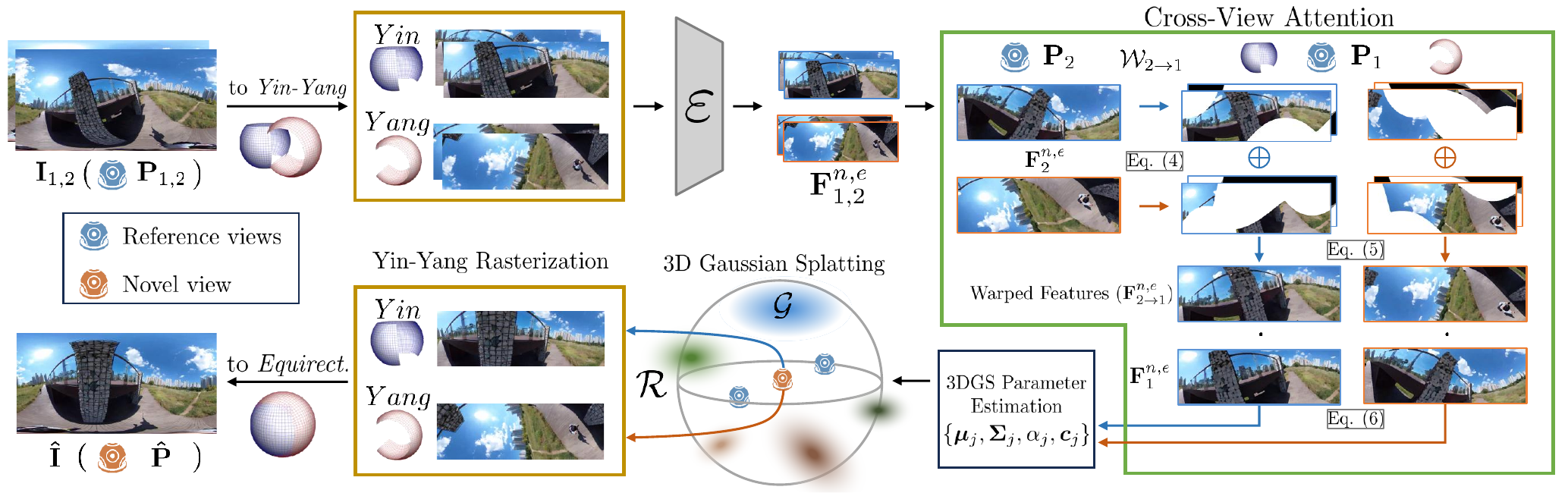}
   \caption{
   \textbf{The overall process of OmniSplat.}
   The two reference omnidirectional images are decomposed into Yin-Yang images, and the cross-view attention is conducted across grids along with epipolar lines to compose cost volume.
   The 3DGS parameters are estimated and Yin-Yang images are rasterized from the novel view.
   The two images are combined to synthesize the final omnidirectional image.
   In cross-view attention, we present red and yellow points and the corresponding sphere sweep curves with the same color. Each image performs cross-attention to the Yin-Yang images from other views, following geometric constraints.
   }
   \label{fig:pipeline}
   \vspace{-3mm}
\end{figure*}

\section{Method}

Omnisplat aims to generate a novel view omnidirectional image by estimating 3D Gaussian splatting from the two reference omnidirectional images and rasterizing the Gaussian splats at a target pose.
Rather than directly training a feed-forward estimation network for omnidirectional images, we propose an effective method to overcome the domain gap between omnidirectional and perspective images by decomposing the omnidirectional images into quasi-uniform Yin-Yang images and exploiting the knowledge of the model trained by a large number of perspective image pairs.

The model receives two reference omnidirectional images and the corresponding poses, which we denote as $\left\{\left( \mathbf{I}_i, \mathbf{P}_i \right)\right\}_{i=1,2}$.
Each image, $\textbf{I}_i$, has $H \times W \times 3$ dimensions, where the width of the image is twice the height.
In~\Cref{sec:crossattention}, we introduce the Yin-Yang grid to decompose an omnidirectional image into two images and conduct the cross-view attention between the two domains.
The small domain gap between Yin-Yang and perspective images encourages the perspective-trained encoder to extract the correct features from the Yin-Yang images.
The Yin-Yang cross-view attention successfully warps the features from a source view to the other view without loss of feature information.
Finally the parameters of 3D Gaussians ($\mathcal{G} = \{ \bm{\mu}_j, \mathbf{\Sigma}_j, \alpha_j, \bm{c}_j \}$) are estimated in a pixel-aligned-manner.
Here, each component indicates the position, covariance, opacity, and spherical harmonics coefficients of j-th Gaussian.
To prevent generating empty regions in the novel-view synthesized images caused by a non-uniform sampling frequency of equirectangular projection, we propose Yin-Yang rasterization to render the omnidirectional image from the estimated 3D Gaussians without producing artifacts.
Instead of rasterizing the omnidirectional image at once, we first rasterize Yin-Yang images from 3DGS and combine the two images in a pixel space.
The detailed process is explained in \Cref{sec:rasterizer}.

\subsection{Yin-Yang Decomposition and Cross-View Attention}
\label{sec:crossattention}

\paragraph{Decomposition based on Yin-Yang grid.}
Although the equirectangular projection provides a one-to-one mapping from the unit sphere to the omnidirectional image, the varying sampling ratio according to the image's latitude makes the characteristics of omnidirectional images different from those of perspective images.
Thus, applying omnidirectional images directly to a model trained with perspective images causes performance degradation due to domain shift.
To circumvent the challenge, we decompose the omnidirectional images into two images using Yin-Yang coordinates~\cite{kageyama2004yin}.
The Yin grid is defined as the following equations:
\begin{equation}
    Yin = \left(\theta \in \left[ -\nicefrac{\pi}{4}, \nicefrac{\pi}{4} \right]\right) \cap \left(\phi \in \left[ -\nicefrac{3\pi}{4}, \nicefrac{3\pi}{4} \right]\right),
\label{eq:yin}
\end{equation}
where $\theta$ and $\phi$ indicate the elevation and azimuth angle of a point on the sphere.
The Yang grid represents the remaining portion of the sphere that is not covered by the Yin grid, which can be computed by rotating the Yin grid using the following transformation matrix $M$.
\begin{equation}
    \begin{pmatrix}
    x_{\text{Yang}} \\ y_{\text{Yang}} \\ z_{\text{Yang}}
    \end{pmatrix}
    = M \begin{pmatrix}
    x_{\text{Yin}} \\ y_{\text{Yin}} \\ z_{\text{Yin}}
    \end{pmatrix}
    = \begin{pmatrix}
    -1 & 0 & 0 \\ 0 & 0 & 1 \\ 0 & 1 & 0
    \end{pmatrix}
    \begin{pmatrix}
    x_{\text{Yin}} \\ y_{\text{Yin}} \\ z_{\text{Yin}}
    \end{pmatrix}.
\label{eq:conv_yang}
\end{equation}

From the omnidirectional image $\mathbf{I}$, we denote the decomposed Yin and Yang images as $\mathbf{I}^n \in \mathbb{R}^{H_n \times W_n \times 3}$ and $\mathbf{I}^e \in \mathbb{R}^{H_e \times W_e \times 3}$ respectively, where $n$ denote the Yin (or north) grid, and $e$ denote the Yang (or east) grid.
We note that $\mathbf{I}^e$s are rotated by 90 degrees from the original Yang-grid images to match the spatial size of $\mathbf{I}^n$s.
Then we concatenate the total four images ($\mathbf{I}_1^n, \mathbf{I}_1^e, \mathbf{I}_2^n, \mathbf{I}_2^e$) to make it look like four view images.

\paragraph{Yin-Yang cross-view attention.}
Following MVSplat~\cite{chen2024mvsplat}, we put four Yin-Yang images into the multiview Transformer to generate features across Yin and Yang images.
We denote the encoded feature as ($\mathbf{F}_1^n, \mathbf{F}_1^e, \mathbf{F}_2^n, \mathbf{F}_2^e$).
\begin{equation}
    (\mathbf{F}_1^n, \mathbf{F}_1^e, \mathbf{F}_2^n, \mathbf{F}_2^e) = \mathbf{\varepsilon} \left( \mathbf{I}_1^n, \mathbf{I}_1^e, \mathbf{I}_2^n, \mathbf{I}_2^e \right).
\end{equation}

We generate the cost volume using a Yin-Yang sweeping approach from the features.
To implement the Yin-Yang sweeping, we define omnidirectional cost volume for each feature, warp the feature from one reference view to the others, and compute the appropriate depth value for each pixel in the feature.
First, we sample a list of depth candidates where the elements compose a harmonic sequence from $d_{near}$ to $d_{far}$, the pre-defined near and far distance values.
According to the value of depth candidates, we warp the feature of one reference view to the pose of the other view.
We assume that we warp the two features, $\mathbf{F}_2^n$ and $\mathbf{F}_2^e$, from pose $\mathbf{P}_2$ (source view) to pose $\mathbf{P}_1$ (target view).
Since we have two features for $\mathbf{I}_1$, the warping should be processed four times to calculate the attention across all domain pairs.
Also, we compute warped masks that indicate the validity of the warping since there are cases where the query point is outside the frustum of the original image when warped.
\begin{equation}
    \mathbf{F}_{2 \rightarrow 1}^{j \rightarrow i}, \mathbf{M}_{2 \rightarrow 1}^{j \rightarrow i} = \mathcal{W}_{2 \rightarrow 1}\left( \mathbf{F}_2^j, \mathbf{P}_1, \mathbf{P}_2\right), \\
\label{eq:cross_warp}
\end{equation}

Here, $i, j \in \{n, e\}$ indicates the grid type (Yin, Yang) of the target and the source features, respectively.
Each warped feature ($\mathbf{F}_{2 \rightarrow 1}$) has shape ${H_F}\times{W_F}\times F \times D$, where $H_F$, $W_F$, $F$ are the height, width, and the number of channels of the feature, and $D$ is the number of depth candidates.
For masks, the value of the mask is set to 1 if the corresponding position is within the image frustum when viewed from $\mathbf{P}_2$, and 0 otherwise.
The final warped feature for each grid is mixed according to the mask:
\begin{equation}
\begin{split}
    \mathbf{F}_{2 \rightarrow 1}^i = \frac{1}{\mathbf{M}^i} (\mathbf{F}_{2 \rightarrow 1}^{n \rightarrow i} \odot \mathbf{M}_{2 \rightarrow 1}^{n \rightarrow i} + \mathbf{F}_{2 \rightarrow 1}^{e \rightarrow i} \odot \mathbf{M}_{2 \rightarrow 1}^{e \rightarrow i}), \\
    \mathbf{M}^i = \mathbf{M}_{2 \rightarrow 1}^{n \rightarrow i}+\mathbf{M}_{2 \rightarrow 1}^{e \rightarrow i},
\end{split}
\label{eq:cross_blend}
\end{equation}
where $\odot$ denotes the pixel-wise multiplication.
From \cref{eq:cross_warp,eq:cross_blend}, Yin and Yang features of each reference view are warped and combined to the novel views in different poses without loss of features.
This eliminates the possibility of not being able to refer to values across grids that can occur with Yin-Yang decomposition.

By calculating the dot product between the warped feature and the corresponding original feature, we calculate the correlation (cross-view attention) and estimate the depth according to the correlation values.
\begin{equation}
    \mathbf{C}_{1}^{i} = \left.\frac{1}{\sqrt{F}} \mathbf{F}_1^{i} \cdot \mathbf{F}_{2 \rightarrow 1}^{i}\right\vert_{i\in\{n,e\}},
    \label{eq:costvolume}
\end{equation}
Here, the inner product is conducted along the channel axis of the feature.
The process of cross-view attention and cost-volume construction is illustrated in~\cref{fig:pipeline}.
After constructing the cost volumes for Yin and Yang images, the properties of 3DGS (position, color, covariance, and opacity) are estimated from the cost volumes, following the estimation network in MVSplat.
The cross-view attention is also conducted from $\mathbf{P}_1$ to $\mathbf{P}_2$, constructing the cost volumes for the second image, $\mathbf{C}_{2}^i$s, where $i$ denotes grid type (\cref{eq:costvolume}).
The estimated Gaussians for all images are unified to produce the final 3D Gaussian splatting.

\subsection{Yin-Yang grid Rasterization}
\label{sec:rasterizer}

It is necessary to rasterize the predicted 3D Gaussian splatting into a 2D image to render the image of the novel view.
There are two challenges for rendering the omnidirectional image from the constructed 3D Gaussians.
First, the rasterizer proposed by~\cite{huang2024error, lee2024odgs} directly renders the omnidirectional images from 3D Gaussian splatting, but it creates artifacts when applied to 3DGS estimated by feed-forward networks.
Second, the unification of 3D Gaussians generated from four images makes a non-uniform 3DGS distribution in the space.
The non-uniformly distributed Gaussians might cause the alpha value of the rasterized image to be inconsistent, which causes unnatural patterns to appear in the image.

We propose Yin-Yang rasterization that utilizes the omnidirectional rasterizer to render Yin-Yang images from 3D Gaussians.
First, Yin image rasterization is processed using the omnidirectional rasterizer  
 while restricting the range as \cref{eq:yin}, denoted as $\mathcal{R}$.
Here, the RGB intensity map and alpha value map are rendered, which we denote as $\hat{V}^n$ and $\hat{A}^n$, respectively.
The rendered image is computed by dividing the intensity map by the alpha value map pixel-wise.
This division removes artifacts where the brightness of the image varies greatly in different regions due to different alpha values.
To render the Yang image, we multiply matrix $M$ in \cref{eq:conv_yang} to the camera rotation matrix and follow the same process in the Yin rasterization.
\begin{equation}
\begin{split}
    \hat{V}^n, \hat{A}^n = \mathcal{R}\left( \mathcal{G}, \mathbf{\hat{P}} \right)&, \mathbf{\hat{I}}^n = \hat{V}^n \odot \nicefrac{1}{\hat{A}^n}, \\
    \hat{V}^e, \hat{A}^e = \mathcal{R}\left( \mathcal{G}, M\mathbf{\hat{P}} \right)&, \mathbf{\hat{I}}^e = \hat{V}^e \odot \nicefrac{1}{\hat{A}^e}.
\end{split}
\end{equation}

Finally, the two images are warped to the omnidirectional grid and combined into a single omnidirectional image.

\begin{table*}[t]
    \centering
    \resizebox{\linewidth}{!}{
    \begin{tabular}{l|c|ccc|ccc|ccc}
        \toprule[1.0pt]
        \multicolumn{1}{c}{Dataset} & & \multicolumn{3}{c|}{OmniBlender} & \multicolumn{3}{c|}{Ricoh360} & \multicolumn{3}{c}{OmniPhotos} \\
        \midrule
        \multicolumn{1}{c|}{Method} & Runtime (s) & PSNR$_{\uparrow}$ & SSIM$_{\uparrow}$ & LPIPS$_{\downarrow}$ & PSNR$_{\uparrow}$ & SSIM$_{\uparrow}$ & LPIPS$_{\downarrow}$ & PSNR$_{\uparrow}$ & SSIM$_{\uparrow}$ & LPIPS$_{\downarrow}$ \\
        \midrule
        PanoGRF~\cite{chen2023panogrf} & 20.110 & 20.45 & \underline{0.6714} & 0.4089 & 17.56 & 0.5827 & 0.4355 & 17.89 & 0.6022 & 0.4270 \\
        PixelSplat (P)~\cite{charatan2024pixelsplat} & 13.068 & 20.56 & 0.6562 & \underline{0.3222} & 19.36 & 0.6307 & \underline{0.3626} & \textbf{19.94} & \underline{0.6486} & \textbf{0.3119} \\
        LatentSplat (P)~\cite{wewer24latentsplat} & 16.097 & 17.28 & 0.5523 & 0.4361 & 16.89 & 0.4113 & 0.5927 & 16.30 & 0.5389 & 0.4386 \\
        MVSplat (P)~\cite{chen2024mvsplat} & 2.224 & 17.48 & 0.5593 & 0.4385 & 18.99 & \textbf{0.6394} & 0.3726 & 18.12 & 0.5947 & 0.3804 \\
        PixelSplat (O)~\cite{charatan2024pixelsplat} & 2.045 & 16.97 & 0.3837 & 0.5967 & 17.06 & 0.3878 & 0.5539 & 16.44 & 0.3546 & 0.5810 \\
        MVSplat (O)~\cite{chen2024mvsplat} & \underline{0.832} & 16.97 & 0.5635 & 0.3949 & 15.68 & 0.4880 & 0.4524 & 15.89 & 0.5478 & 0.4049 \\ 
        DepthSplat (O)~\cite{xu2024depthsplat} & 1.602 & 17.89 & 0.5364 & 0.4753 & 15.85 & 0.5012 & 0.5100 & 17.43 & 0.5470 & 0.4482 \\
        MVSplat (Y)~\cite{chen2024mvsplat} & 2.096 & 18.43 & 0.5627 & 0.4107 & 17.88 & 0.5482 & 0.4123 & 17.76 & 0.5768 & 0.4051 \\ 
        OmniSplat & \textbf{0.532} & \underline{21.22} & 0.6519 & 0.3636 & \underline{19.72} & 0.6041 & 0.3886 & 18.59 & 0.6195 & 0.4115 \\
        OmniSplat+\textit{opt} & 12.04 & \textbf{22.47} & \textbf{0.6960} & \textbf{0.3209} & \textbf{20.63} & \underline{0.6367} & \textbf{0.3541} & \underline{19.57} & \textbf{0.6562} & \underline{0.3544} \\
        \midrule
        ODGS~\cite{lee2024odgs} & 1920 & 22.23& 0.6807 & 0.3124 & 17.51 & 0.5309 & 0.3911 & 20.25 & 0.5660 & 0.3730 \\
        \midrule
        \midrule
        \multicolumn{1}{c}{Dataset} & & \multicolumn{3}{c|}{360Roam} & \multicolumn{3}{c|}{OmniScenes} & \multicolumn{3}{c}{360VO} \\
        \midrule
        \multicolumn{1}{c}{Method} & & PSNR$_{\uparrow}$ & SSIM$_{\uparrow}$ & LPIPS$_{\downarrow}$ & PSNR$_{\uparrow}$ & SSIM$_{\uparrow}$ & LPIPS$_{\downarrow}$ & PSNR$_{\uparrow}$ & SSIM$_{\uparrow}$ & LPIPS$_{\downarrow}$ \\
        \midrule
        \multicolumn{1}{l}{PanoGRF~\cite{chen2023panogrf}} & & \underline{17.87} & \textbf{0.5598} & \underline{0.4621} & 20.14 & 0.7579 & 0.3530 & 20.52 & 0.7077 & \underline{0.2618} \\
        \multicolumn{1}{l}{PixelSplat (P)~\cite{charatan2024pixelsplat}} & & 16.31 & 0.5270 & \textbf{0.4415} & 19.88 & 0.7193 & 0.3502 & 19.33 & 0.6628 & 0.3079 \\
        \multicolumn{1}{l}{LatentSplat (P)~\cite{wewer24latentsplat}} & & 15.58 & 0.5315 & 0.4731 & 17.04 & 0.6907 & 0.3998 & 18.36 & 0.6457 & 0.3345 \\
        \multicolumn{1}{l}{MVSplat (P)~\cite{chen2024mvsplat}} & & 15.96 & 0.5249 & 0.4930 & 19.06 & 0.6954 & 0.3710 & 19.19 & 0.6322 & 0.3118 \\
        \multicolumn{1}{l}{PixelSplat (O)~\cite{charatan2024pixelsplat}} & & 14.87 & 0.2614 & 0.6430 & 17.58 & 0.4152 & 0.6082 & 17.79 & 0.4635 & 0.5411 \\
        \multicolumn{1}{l}{MVSplat (O)~\cite{chen2024mvsplat}} & & 12.58 & 0.3680 & 0.5221 & 13.68 & 0.7202 & 0.3635 & 17.45 & 0.6695 & 0.2991 \\
        \multicolumn{1}{l}{DepthSplat (O)~\cite{xu2024depthsplat}} & & 16.00 & 0.4633 & 0.5390 & 18.30 & 0.6772 & 0.4427 & 18.69 & 0.6170 & 0.3641 \\
        \multicolumn{1}{l}{MVSplat (Y)~\cite{chen2024mvsplat}} & & 16.54 & 0.4859 & 0.5044 & 20.18 & 0.7195 & 0.3659 & 19.85 & 0.6922 & 0.2730 \\
        \multicolumn{1}{l}{OmniSplat} & & 17.53 & 0.5264 & 0.5170 & \underline{21.59} & \underline{0.7649} & \underline{0.3386} & \underline{20.63} & \underline{0.7167} & 0.2859 \\
        \multicolumn{1}{l}{OmniSplat+\textit{opt}} & & \textbf{18.19} & \underline{0.5569} & 0.4704 & \textbf{23.12} & \textbf{0.7985} & \textbf{0.2939} & \textbf{21.42} & \textbf{0.7485} & \textbf{0.2506} \\
        \midrule
        \multicolumn{1}{l}{ODGS~\cite{lee2024odgs}} & & 18.72 & 0.5630 & 0.3833 & 20.71 & 0.7598 & 0.2749 & 22.66 & 0.7786 & 0.2222 \\
        \bottomrule
    \end{tabular}
    }
    \caption{
    \textbf{Quantitative comparison.} We list the performance metrics of reconstructed results in novel view omnidirectional images with existing feed-forward networks and an optimization-based approach on various datasets.
    The best and the second best scores among feed-forward methods are written in \textbf{bold} and \underline{underlined}, respectively.
    }
    \vspace{-2mm}
    \label{tab:comp_ego}
\end{table*}

\section{Experiments}

\paragraph{Datasets.}
We use six benchmark omnidirectional videos for comparing the reconstruction quality.
OmniBlender~\cite{choi2023balanced} consists of 11 synthetic videos where each 3D scene is generated and rendered by Blender engine~\cite{blender}.
We use 25 test images, which are uniformly sampled from each video for each scene.
Ricoh360~\cite{choi2023balanced} contains 12 videos captured in outdoor scenes using real 360-degree cameras.
Each scene has 50 test images that are used for comparing our method with other baselines.
OmniPhotos~\cite{bertel2020omniphotos} captured 10 real-world outdoor videos by rotating the 360-degree camera in a circle.
Each scene consists of 71 to 91 images, and we uniformly sample 20\% of images as test views for the comparison.
While the aforementioned three datasets have small camera displacement between frames, we also compare our model with the other three benchmarks with large motions: 360Roam~\cite{huang2022tc360roam}, OmniScenes~\cite{kim2021piccolo}, and 360VO~\cite{huang2022360vo}.

For all datasets, we run OpenMVG~\cite{moulon2016openmvg} to obtain a pose matrix for each image since there is no pose given in the dataset.
Also, we downsample all images into $1024 \times 512$ using bicubic downsampling for all scenes.

\noindent\textbf{Metrics.}
For evaluation, we use three metrics to compare the quality of test view images: PSNR, SSIM, and LPIPS.
PSNR and SSIM~\cite{measure_ssim} are popularly used metrics to indicate restoration accuracy.
LPIPS~\cite{feat_deep} is a well-known metric to measure the perceptual quality of the output compared to the ground truth.
We use VGG~\cite{net_vgg} backbone for computing the distance between the features when calculating LPIPS.

\noindent\textbf{Experimental details.}
Since no datasets have been proposed for 3D reconstruction from sparse inputs, we set two new image indices to be used as reference views.
For OmniBlener, Ricoh360, and OmniPhotos, we select two indices as reference views for each scene where the two images are taken from a distance to minimize the portion of occluded regions.
On the contrary, for the other three datasets (360Roam, OmniScenes, and 360VO), the camera movement distance between adjacent frames is long, so the overlap ratio between frames is low. Therefore, we set two frames with an interval of 4 to 5 timesteps as reference images and designated the frames in between as test views to compare the restoration performance.
The detailed indices for the reference views and test views are written in the supplementary materials.
We manually set $d_{near}$ and $d_{far}$ to 1 and 100
The parameters of the feed-forward 3DGS estimation part of OmniSplat are loaded from the pre-trained model of MVSplat~\cite{chen2024mvsplat}.

\subsection{Novel View Synthesis}
\label{sec:nvs}
To the best of our knowledge, no works estimate the whole 3D Gaussian splatting from a few multiview omnidirectional images.
Thus, we changed the dataset format and transformed some functions of the model to correspond to omnidirectional images and then compared them with OmniSplat.
\Cref{tab:comp_ego} shows the quantitative results for the test view images according to the synthesis methods.

\begin{figure*}[t]
    \newcommand{\ww}{0.195\linewidth}
    \newcommand{\hh}{0.15\linewidth}
    \centering

    \addtocounter{subfigure}{-5}
    \subfloat{\includegraphics[width=\ww,height=\hh]{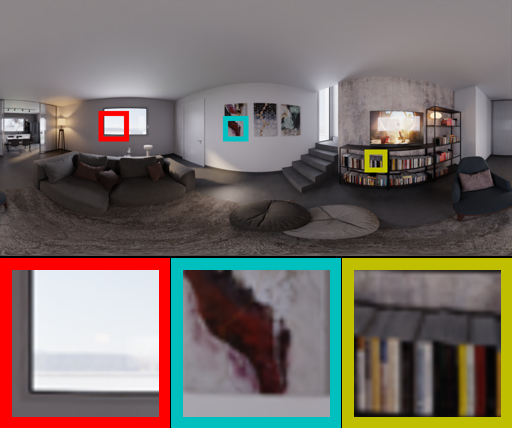}}
    \hfill
    \subfloat{\includegraphics[width=\ww,height=\hh]{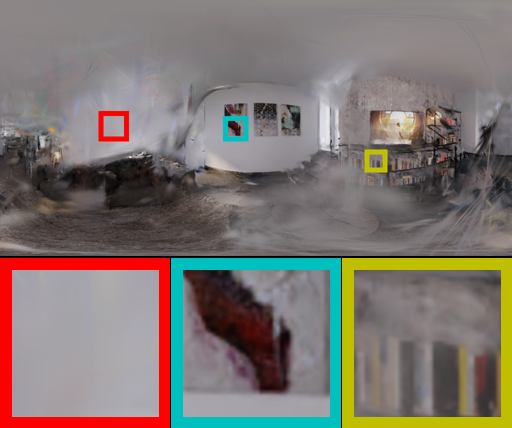}}
    \hfill
    \subfloat{\includegraphics[width=\ww,height=\hh]{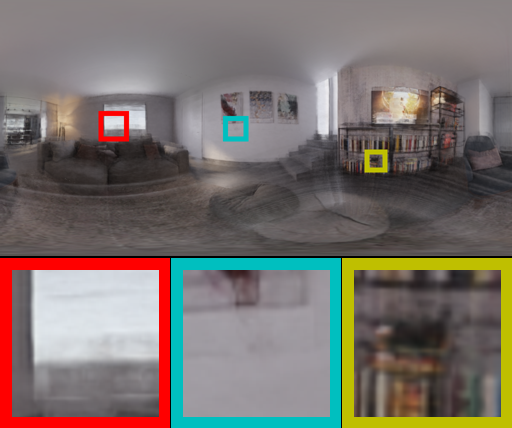}}
    \hfill
    \subfloat{\includegraphics[width=\ww,height=\hh]{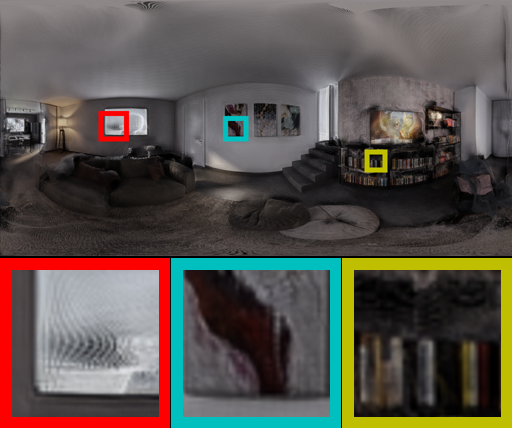}}
    \hfill
    \subfloat{\includegraphics[width=\ww,height=\hh]{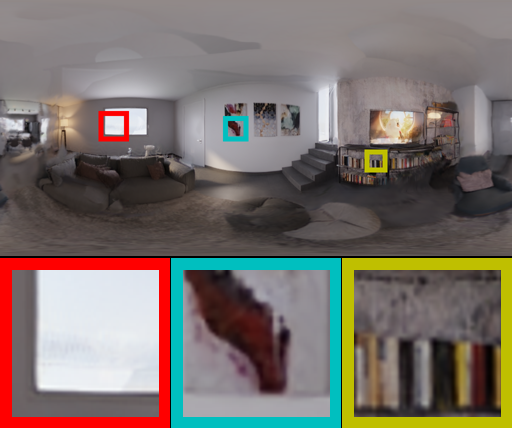}}
    \vspace{0.5mm}
    
    \addtocounter{subfigure}{-5}
    \subfloat{\includegraphics[width=\ww,height=\hh]{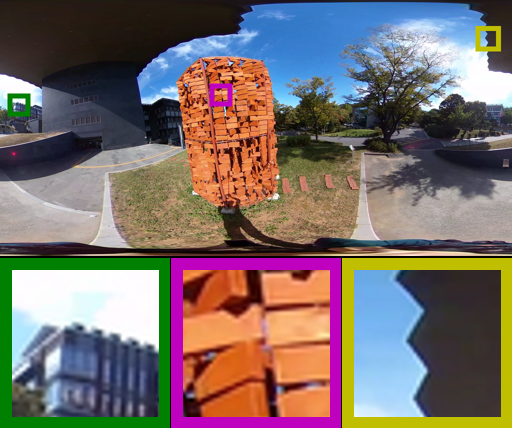}}
    \hfill
    \subfloat{\includegraphics[width=\ww,height=\hh]{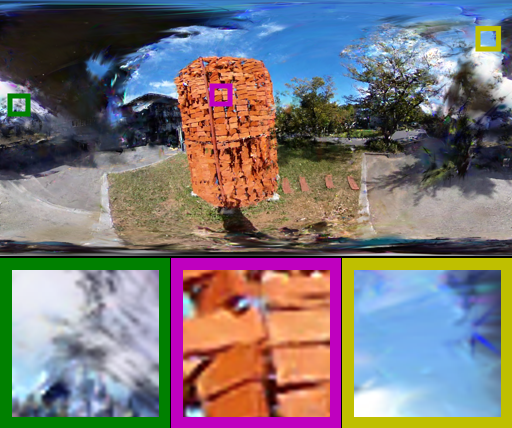}}
    \hfill
    \subfloat{\includegraphics[width=\ww,height=\hh]{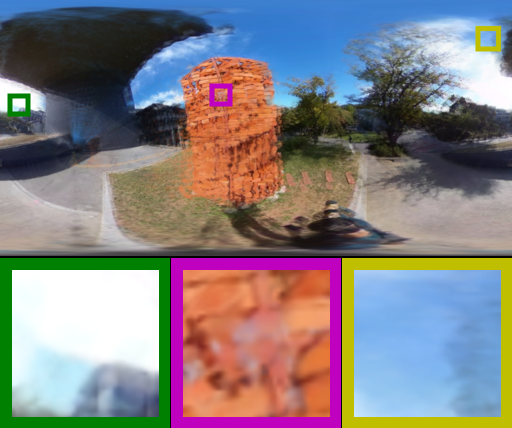}}
    \hfill
    \subfloat{\includegraphics[width=\ww,height=\hh]{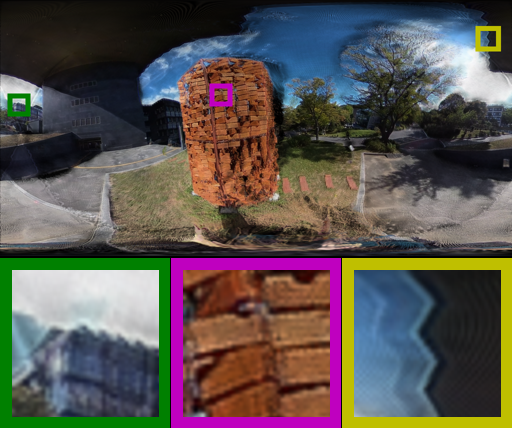}}
    \hfill
    \subfloat{\includegraphics[width=\ww,height=\hh]{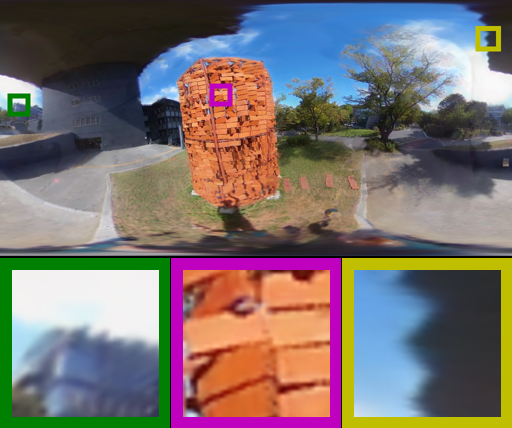}}
    \vspace{0.5mm}

    \addtocounter{subfigure}{-5}
    \subfloat[Ground truth]{\includegraphics[width=\ww,height=\hh]{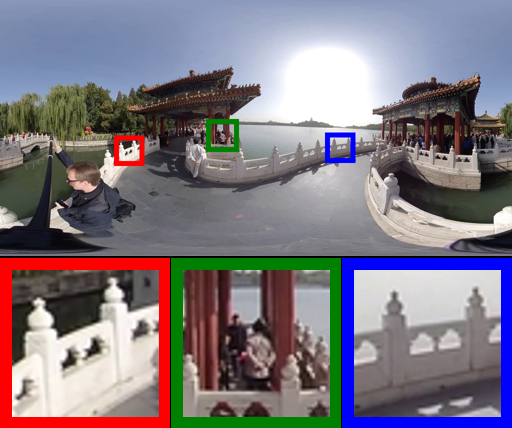}}
    \hfill
    \subfloat[ODGS~\cite{lee2024odgs}]{\includegraphics[width=\ww,height=\hh]{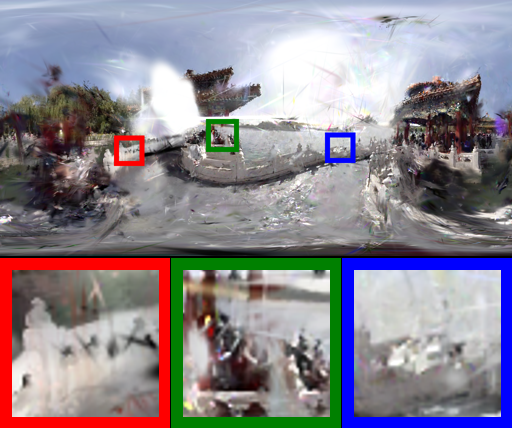}}
    \hfill
    \subfloat[PixelSplat (P)~\cite{charatan2024pixelsplat}]{\includegraphics[width=\ww,height=\hh]{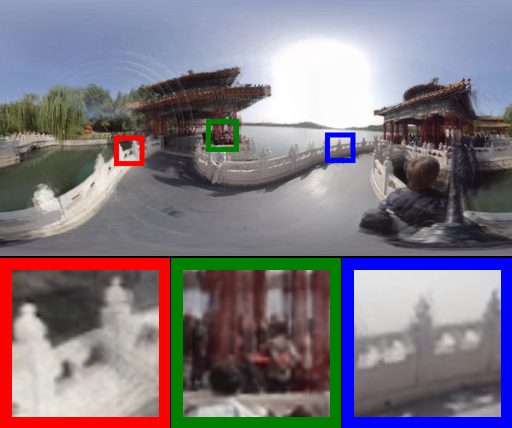}}
    \hfill
    \subfloat[MVSplat (O)~\cite{chen2024mvsplat}]{\includegraphics[width=\ww,height=\hh]{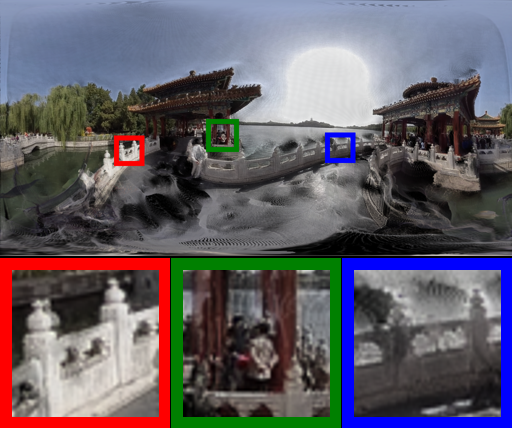}}
    \hfill
    \subfloat[OmniSplat]{\includegraphics[width=\ww,height=\hh]{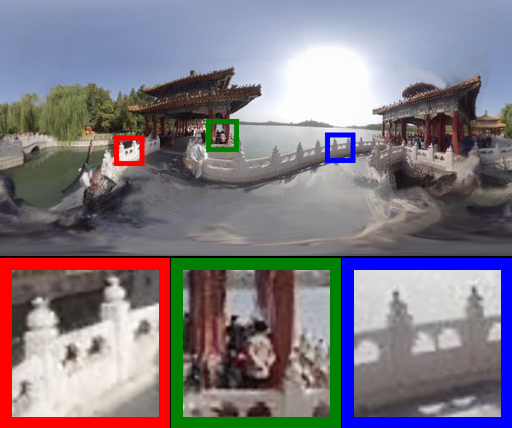}}
    
    \caption{\textbf{Qualitative comparison.} Novel view synthesized image examples in various datasets.
    Each scene is brought from OmniBlender, Ricoh360, and OmniPhotos, respectively.
    \textit{Best viewed when zoomed in.}
    }
    \label{fig:qual_ego}
    \vspace{-1mm}
\end{figure*}

First, PixelSplat~\cite{charatan2024pixelsplat}, LatentSplat~\cite{wewer24latentsplat}, and MVSplat~\cite{chen2024mvsplat} are designed to generate parameters of 3D Gaussian splatting from a few perspective images.
However, their models only support perspective images and cannot be used for omnidirectional images.
To use those models as baselines, we decompose each input omnidirectional image into six perspective images using cubemap decomposition.
Cubemap decomposition draws a virtual cube that touches the unit sphere and then creates six perspective images and the corresponding poses where each face is considered a camera plane.
Then, 12 perspective images are put into the network, generating six perspective images for the test camera pose, and they are combined with the omnidirectional image.
Since PixelSplat and LatentSplat cannot receive more than two images, we choose one image per view, construct a set of 36 pairs of products, and unify all Gaussian splats for all pairs.
Then, the omnidirectional image is synthesized from six rasterized cubemap images.
For those cases, we write `(P)' after the model names in the method column to indicate that the model is executed with the perspective images.
Compared to OmniSplat, perspective feed-forward models take a much longer time to generate the output.
This is because it takes a long time to compute cross-attention between 12 images or predict Gaussians of 36 pairs, and it also takes time to warp six cubemap images to stitch them.
Reconstruction performance is measured lower than OmniSplat except for PixelSplat, as features were mixed with too many views or too many Gaussians were unified.

Meanwhile, we modify some functions of PixelSplat, MVSplat, and DepthSplat~\cite{xu2024depthsplat}, which calculate the ray direction according to the projection on the image pixel or the camera plane, to fit the optical characteristics of equirectangular projection so that they can be applied to omnidirectional images.
After creating 3DGS, we directly synthesize novel view omnidirectional images using omnidirectional rasterizer~\cite{lee2024odgs}.
The character `(O)' after the model name indicates that the omnidirectional version of the model is utilized.
Although these models showed faster generation speeds than the perspective versions, the quality of the generated results was measured to be worse.
We speculate that this happens because the feature extractor trained on perspective images fails to obtain appropriate features with omnidirectional images and because of the dense sampling in the polar regions of the equirectangular projection.

\begin{table*}[ht]
    \centering
    \begin{minipage}[t]{0.45\linewidth}
    \resizebox{\linewidth}{!}{
    \begin{tabular}{cc|ccc}
        \toprule
        Encoder & Rasterizer & PSNR$_{\uparrow}$ & SSIM$_{\uparrow}$ & LPIPS$_{\downarrow}$ \\
        \midrule
        Yin-Yang & Omnidirectional & 13.30 & 0.4836 & 0.4965 \\
        Omnidirectional & Yin-Yang & 15.97 & 0.5717 & \textbf{0.4001} \\
        \midrule
        Yin-Yang & Yin-Yang & \textbf{18.59} & \textbf{0.6195} & 0.4115 \\
        \bottomrule
    \end{tabular}
    }
    \subcaption{OmniPhotos~\cite{bertel2020omniphotos}}
    \end{minipage}
    \hspace{0.06\linewidth}
    \begin{minipage}[t]{0.45\linewidth}
    \resizebox{\linewidth}{!}{
    \begin{tabular}{cc|ccc}
        \toprule
        Encoder & Rasterizer & PSNR$_{\uparrow}$ & SSIM$_{\uparrow}$ & LPIPS$_{\downarrow}$ \\
        \midrule
        Yin-Yang & Omnidirectional & 16.59 & 0.6466 & 0.4462 \\
        Omnidirectional & Yin-Yang & 17.82 & 0.7086 & 0.3726 \\
        \midrule
        Yin-Yang & Yin-Yang & \textbf{21.44} & \textbf{0.7649} & \textbf{0.3386} \\
        \bottomrule
    \end{tabular}
    }
    \subcaption{OmniScenes~\cite{kim2021piccolo}}
    \end{minipage}
    \caption{\textbf{Effect of Yin-Yang decomposition for attention and rasterization.}
    We change the image domain for attention in the encoder and rasterization.
    The best metric is written in \textbf{bold}.
    }
    \label{tab:ablation_yinyang}
    \vspace{-3mm}
\end{table*}

Since MVSplat can receive more than two images, we directly feed four Yin-Yang decomposed images to MVSplat, construct the 3D Gaussian splats, and write the performance of rendered novel-view images in the row `MVSplat (Y).'
We note that the proposed Yin-Yang rasterization process is utilized after the Gaussian splats are reconstructed.
The na\"ive combination of Yin-Yang decomposition and MVSplat yields inferior results since the warped features from the other three images are pixel-wise averaged during cost volume construction for a query view.
The pixel-wise averaging leads to an incorrect cost volume estimation and novel-view synthesis.
The results validate that OmniSplat produces much more accurate results than MVSplat with Yin-Yang decomposed inputs.

ODGS~\cite{lee2024odgs} is an optimization-based 3DGS estimation method specifically designed for omnidirectional image input.
In the experiment, we measure the performance after 30,000 iterations of optimization for each scene, which takes about 30 minutes.
Even though Gaussians are optimized for each scene, the results do not outperform than the feed-forward networks and even show worse metrics in some datasets.
This phenomenon occurs because overfitting often occurs during optimization with only two reference images; ODGS shows the best results in reference views, but the image is completely broken when rendered in novel views. 

OmniSplat shows the best results for most settings, demonstrating the power of the Yin-Yang decomposition and the corresponding rasterizer with the fastest rendering speed.
Furthermore, we present `OmniSplat+\textit{opt}', after a small amount of optimization of the color and opacity properties in 3DGS with reference views.
Different from ODGS, our method is much more robust to overfitting since 3DGS predicted by OmniSplat serve as good initialization points, and their positions are not updated during optimization.
While reporting similar runtime to PixelSplat~(P), OmniSplat+\textit{opt} shows better performance in the majority of datasets.

The novel-view synthesized images according to 3DGS reconstruction methods on various datasets are illustrated in~\Cref{fig:qual_ego}.
Despite reporting decent metrics, images synthesized by ODGS contain many artifacts since the Gaussians are overfitted to the reference views, generating wrong results in novel viewpoints.
PixelSplat~(P), despite reporting comparable quantitative performance, produces misaligned and blurry images due to the overlapping of Gaussians made from many perspective pairs.
Additionally, if the image only has simple textures, such as the sky and walls, it may not be able to create meaningful features, leading to incorrect depth estimation and inaccurate restoration results.
In the case of MVSplat (O), the overall tone of the photo appears darker than the ground truth.
Also, stripe-shaped artifact patterns occur frequently at the top and bottom of the image.
This artifact occurs because the Gaussians located in 3D space are not dense or large enough to produce an accurate pattern when rasterized with the high sampling rate at the pole regions.
Consequently, some pixels appear bright as they pass through a Gaussian, while other areas appear dark due to insufficient coverage.
Our method, OmniSplat, generates the most accurate and photo-realistic images compared to previous methods.

\paragraph{Effects of cross-view attention and rasterization.}
We conducted the experiment and analyzed the effect of Yin-Yang cross-view attention on encoding and Yin-Yang rasterizer on decoding in \Cref{tab:ablation_yinyang}.
First, replacing the Yin-Yang rasterizer with the omnidirectional rasterizer results in a significant performance drop since the generated omnidirectional images include severe artifacts with empty holes caused by the non-uniform sampling.
Although better than the first combination, using an omnidirectional image pair to construct cost volume with Yin-Yang rasterization produces inferior results.
The Yin-Yang rasterizer prevents artifact generation in the image, resulting in better synthesis quality than the previous configuration.
However, the domain gap between the omnidirectional and perspective makes the encoder fail to extract appropriate features, leading to the incorrect estimation of 3D Gaussians.
OmniSplat shows the best PSNR and SSIM, demonstrating the advantage of Yin-Yang decomposition in both encoder and rasterizer.

\subsection{Multiview Consistent Segmentation}
\begin{figure}[t]
    \newcommand{\ww}{0.496\linewidth}
    \newcommand{\hh}{0.248\linewidth}
    \centering

    \subfloat{\includegraphics[width=\ww, height=\hh]{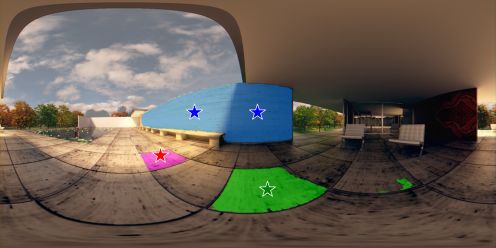}}
    \hfill
    \subfloat{\includegraphics[width=\ww, height=\hh]{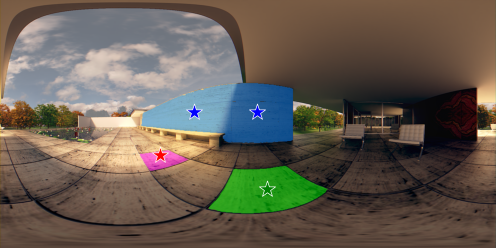}}

    \addtocounter{subfigure}{-2}
    \subfloat[DEVA~\cite{cheng2023deva}]{\includegraphics[width=\ww, height=\hh]{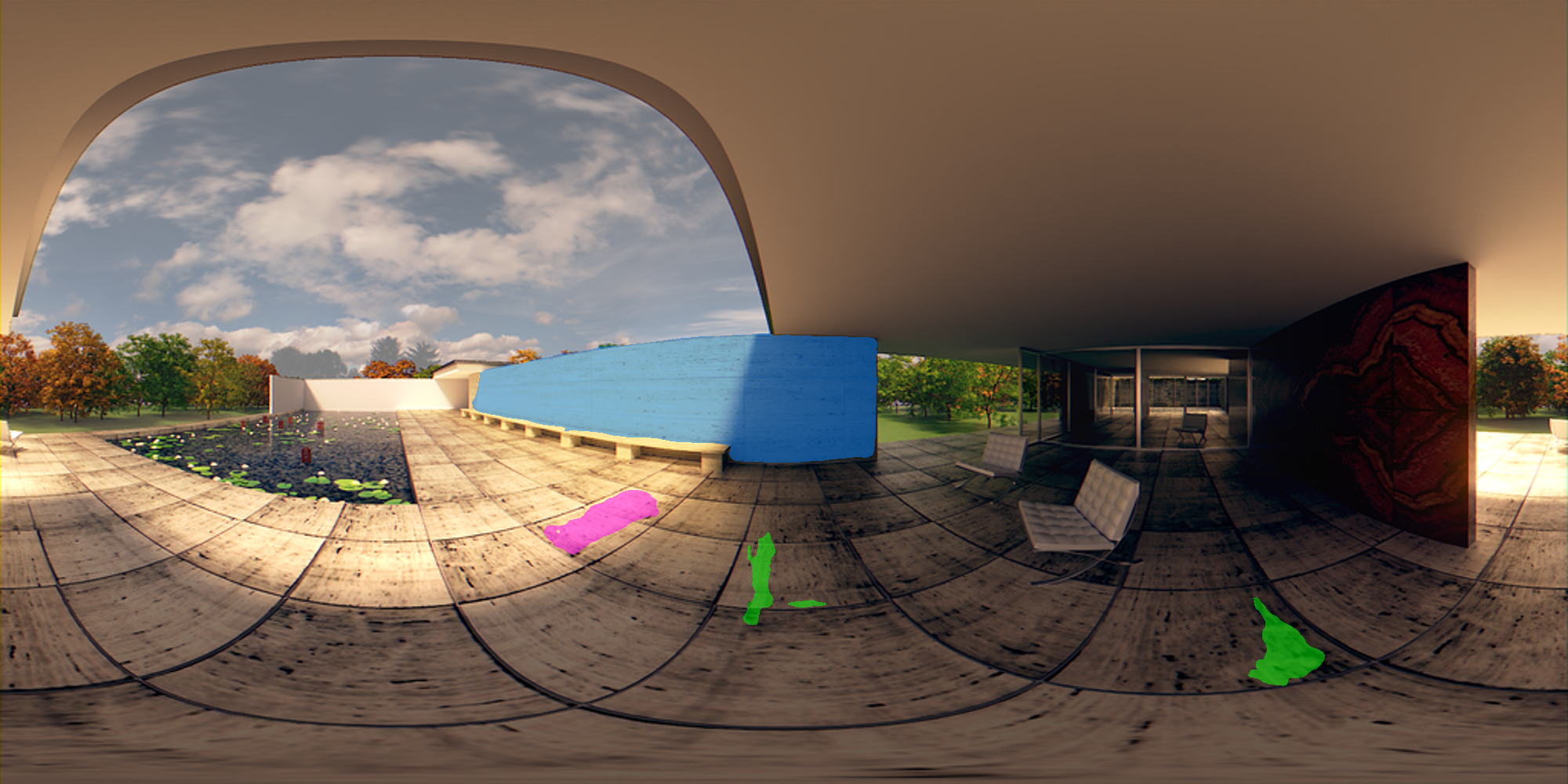}\label{fig:qual_editing_a}} 
    \hfill
    \subfloat[OmniSplat]{\includegraphics[width=\ww, height=\hh]{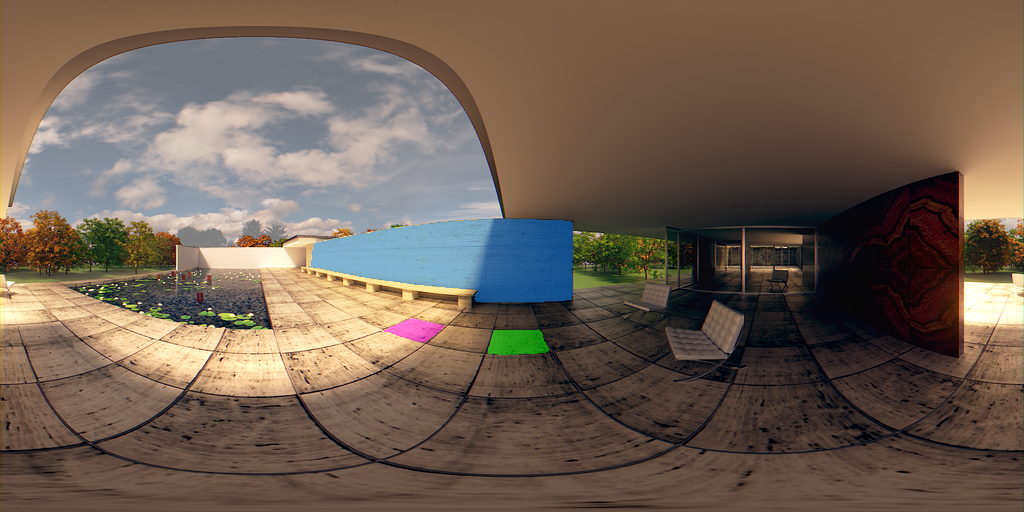}\label{fig:qual_editing_b}}
    
    \caption{
        \textbf{Visualization of segment matching.} We visualize the matched segment samples among the source (top) and the target (bottom) views.
        The stars in the image indicate query points for the user to segment objects containing the stars.
    }
    \label{fig:qual_editing}
\end{figure}
\begin{table}[t]
    \centering
    \resizebox{\linewidth}{!}{
    \begin{tabular}{c|cccccc}
        \toprule
         Dataset & OmniBlender & Ricoh360 & OmniPhotos & 360Roam & OmniScenes & 360VO \\
         \midrule
         DEVA~\cite{cheng2023deva} & 0.5354 & 0.5969 & 0.7034 & 0.6657 & \textbf{0.8200} & 0.7951 \\
         OmniSplat & \textbf{0.5590} & \textbf{0.6195} & \textbf{0.7280} & \textbf{0.6921} & 0.8181 & \textbf{0.8098} \\
         \bottomrule
    \end{tabular}
    }
    \caption{\textbf{Segmentation matching accuracy (mIoU) comparison.}}
    \label{tab:seg}
\end{table}
We obtain multiview consistent segments by leveraging attention-based matching across views.
This enables efficient, feed-forward 3D segmentation and editing without the process of explicit grouping in 3D space. 
We establish a matching between estimated segments across the images based on the attention scores, thereby aligning the labels and ensuring consistency. 
Each pixel in the reference view matches the most focused pixel in the other view based on cross-attention scores. 
We compare our attention-based matching approach with the segmentation tracking method DEVA~\cite{cheng2023deva} as illustrated in \Cref{fig:qual_editing}. 
In the top row, the selected points from the source view are marked with stars of different colors.
Regions sharing the same label as the selected points are highlighted with corresponding color masks. 
The second row of \Cref{fig:qual_editing} shows how these regions correspond in the target view with the tracking-based method and our attention-based method. 
DEVA, a video tracking-based method in \Cref{fig:qual_editing_a}, shows some tracking failures when the camera moves significantly, resulting in the disappearance or mixing of regions identified in the source view. 
In contrast, our approach demonstrated in \Cref{fig:qual_editing_b} achieves multi-view consistent segmentation. 
Notably, we identify matched segments based on attention scores obtained during the reconstruction process, which enables the retrieval of corresponding segmentation regions without additional computation.
In \Cref{tab:seg}, our approach achieves superior segmentation correspondence, enabling multiview consistent editing.
Due to the absence of ground truth segment maps in the datasets, we applied DEVA tracking over full image sequences and treated the result as pseudo ground truth for evaluation.
These view-consistent segments also allow us to select semantic Gaussians for downstream tasks such as 3D editing.
We demonstrate other examples of Gaussian segmentation and editing in the supplementary material.

\section{Conclusion}

In this work, we propose OmniSplat, a pioneering work for a feed-forward 3D scene reconstruction from omnidirectional images.
Although omnidirectional images have a wide field-of-view that can capture the entire scene in a single frame, existing feed-forward models~\cite{charatan2024pixelsplat,chen2024mvsplat,wewer24latentsplat} cannot be used with omnidirectional images due to the difference of optical characteristics from perspective images.
Specifically, optical distortion at the edges of an omnidirectional image can force the model to extract incorrect features, and non-uniformly generated Gaussians often create artifacts during rendering.
To cope with the issues, we introduce Yin-Yang grid, which can represent spheres in a quasi-uniform manner while having a typical lattice structure.
The two decomposed images and the corresponding features are warped with cross-view attention to estimate the depth of each pixel and parameters of 3DGS.
The proposed Yin-Yang rasterizer successfully renders the Yin-Yang images at the query view, which are combined in pixel space to synthesize the omnidirectional image.
OmniSplat shows more accurate and visually pleasing reconstruction results than previous feed-forward networks designed for perspective images while achieving faster inference speed.

\noindent\textbf{Limitations and future works.}
Although OmniSplat shows superiority in rendering novel view omnidirectional images without fine-tuning the existing MVSplat model, there is some room for further improvement.
First, when transforming a Yang image to an omnidirectional domain, interpolation is required in the horizontal direction, reducing the quality of the image.
Also, we believe that the adaptation steps can be further reduced through meta-learning.
\clearpage
\section*{Acknowledgements}
This work was supported in part by the IITP grants [No.2021-0-01343, Artificial Intelligence Graduate School Program (Seoul National University), No. 2021-0-02068, and  No.2023-0-00156], the NOTIE grant (No. RS-2024-00432410) by the Korean government, and the SNU-LG AI Research Center.

{
    \small
    \bibliographystyle{ieeenat_fullname}
    \bibliography{main}
}

\clearpage
\maketitlesupplementary

\setcounter{section}{0}
\setcounter{figure}{0}
\setcounter{table}{0}

\renewcommand{\thesection}{\Alph{section}}
\renewcommand{\thetable}{\Alph{table}}
\renewcommand{\thefigure}{\Alph{figure}}

\section{Details on OmnniSplat+opt}

As briefly mentioned in \Cref{sec:nvs}, we add a small number of optimizations for each scene using the reference view images for fast performance improvement and call the method `OmniSplat+\textit{opt}.'
Compared to ODGS~\cite{lee2024odgs}, the optimization process of OmniSplat+\textit{opt} is different in many aspects.
First, while ODGS starts optimization from the sparse point cloud estimated by OpenMVG~\cite{moulon2016openmvg}, our model starts from 3D Gaussians generated by OmniSplat, which contains much more information.
Next, we set the number of optimization steps of OmniSplat+\textit{opt} to 100, whereas ODGS optimizes 30,000 times.
Since OmniSplat's initial points are highly accurate, significant performance improvement can be achieved with a small number of iterations.
Thus, the optimization time only takes 11 seconds, which is more than 150 times shorter than ODGS (32 minutes).
Finally, we only optimize the color (sh coefficients) and opacity properties of 3DGS and keep the position and covariance fixed since changes in position or covariance can cause overfitting when optimizing with a few images.

\paragraph{Ablation Studies: number of optimizing iterations}
We measure the performance of test view images according to the number of optimizing iterations, starting from the original OmniSplat.
As mentioned, optimization with reference images improves performance in test view images in the earlier stage.
However, if we optimize 3DGS for a long time only with the reference views, the 3DGS may be overfitted, and the performance of the test views may be saturated or even reduced.
\Cref{tab:ab_adapt} shows the changes of metrics (PSNR, SSIM, and LPIPS) as well as optimization time according to the number of optimization iterations (\# opt.) in OmniBlender~\cite{choi2023balanced}.
PSNR grows rapidly at the first 100 iterations but becomes saturated as optimization proceeds.
For SSIM and LPIPS, the best value is achieved after 200 and 400 iterations of optimization, and gets worse when the number of iterations gets larger.
In terms of time, the optimization takes about 12 seconds for 100 iterations, and the time linearly grows as the number of optimizations increases.
We find out that 100 or 200 iterations are optimal for additional optimization from OmniSplat, considering both time and performance.
In the main manuscript, we use 100 times optimization, which takes a similar time to PixelSplat or LatentSplat.

\begin{figure}[t]
   \centering
   \includegraphics[width=1\linewidth]{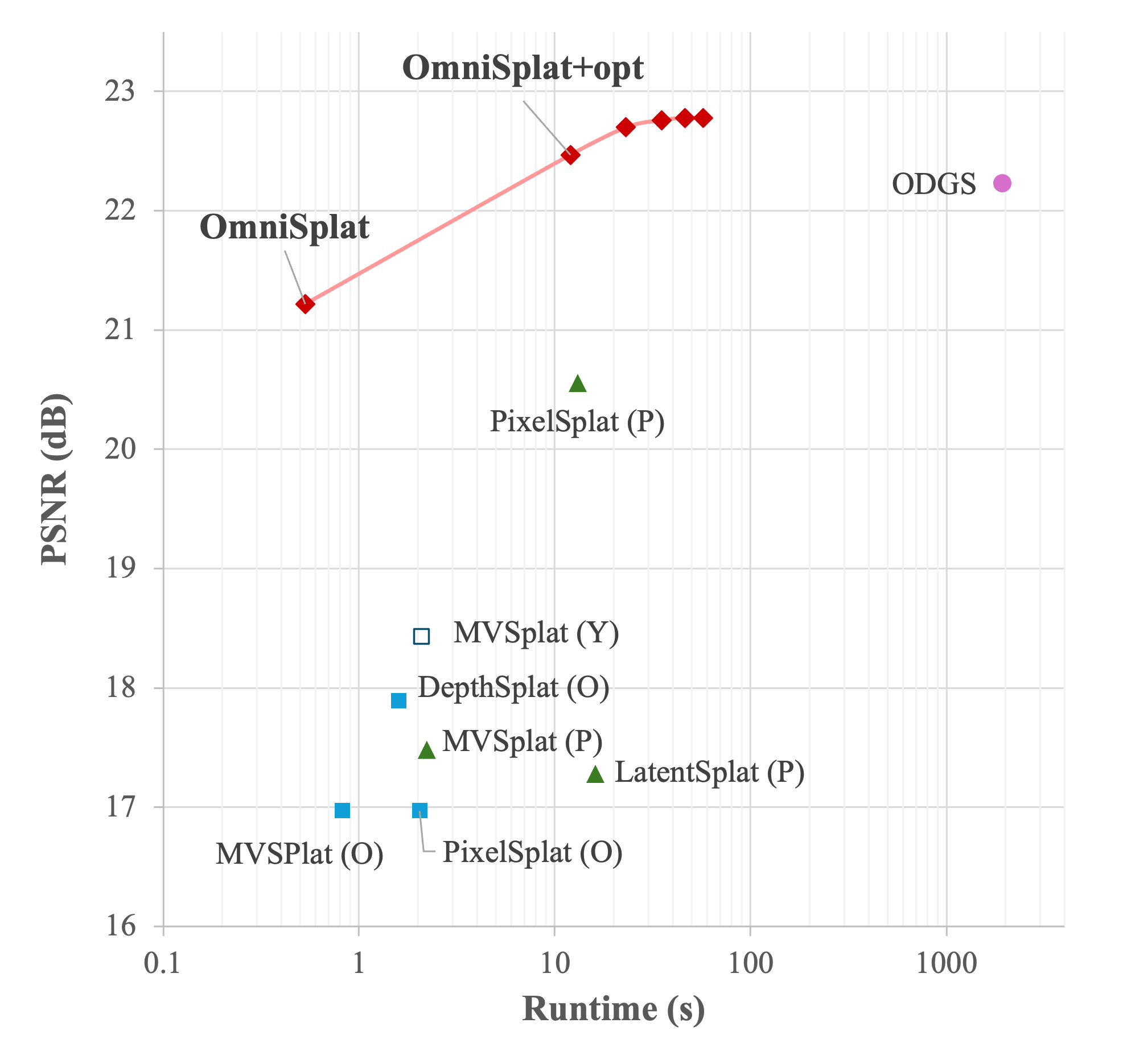}
   \caption{
        \textbf{PSNR-runtime trade-off.} A chart of reconstruction performance-runtime trade-off in novel view image on OmniBlender~\cite{choi2023balanced}, including the ablation according to the number of optimizations.
   }
   \label{fig:supp_tradeoff}
\end{figure}

We illustrate the PSNR-runtime trade-off, including the values of \Cref{tab:ab_adapt} in \Cref{fig:supp_tradeoff}.
Compared to existing feed-forward models, OmniSplat achieves higher PSNR with the fastest running speed.
Additionally, the PSNR values get higher than ODGS within a minute of optimization, which are denoted as OmniSplat+\textit{opt}.
From the figure, we maintain that OmniSplat achieves the best PSNR-runtime trade-off for novel view synthesis.

\begin{table}[t]
    \centering
    \resizebox{\linewidth}{!}{
    \begin{tabular}{c|cccccc}
        \toprule[1.0pt]
        \# opt. & 0 & 100 & 200 & 300 & 400 & 500 \\
        \midrule
        PSNR &  21.22  & 22.47  & 22.70  & 22.76  & 22.78  & \textbf{22.78} \\
        SSIM &  0.6519 & 0.6960 & \textbf{0.6979} & 0.6962 & 0.6938 & 0.6914 \\
        LPIPS & 0.3636 & 0.3209 & 0.3144 & 0.3125 & \textbf{0.3124} & 0.3129 \\
        \midrule
        Time (s) & 0 & 12 & 23 & 35 & 46 & 57 \\
        \bottomrule
    \end{tabular}
    }
    \caption{
        Changes in performance according to the number of iterations for adaptation in OmniBlender~\cite{choi2023balanced}.
    }
    \label{tab:ab_adapt}
\end{table}
\begin{figure*}[ht]
    \newcommand{\ww}{0.246\linewidth}
    \newcommand{\hh}{0.123\linewidth}
    \centering

    \addtocounter{subfigure}{-4}
    \subfloat{\includegraphics[width=\ww, height=\hh]{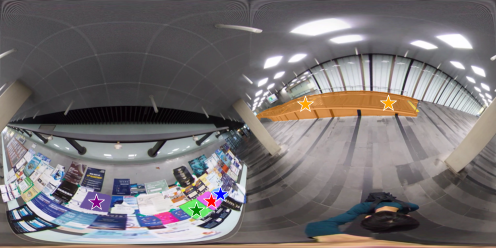}}
    \hfill
    \subfloat{\includegraphics[width=\ww, height=\hh]{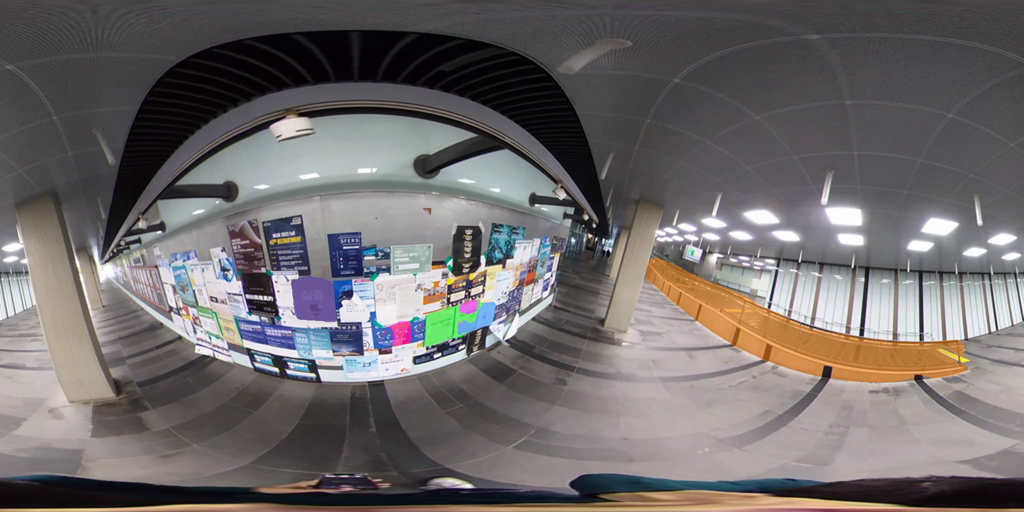}}
    \hfill
    \subfloat{\includegraphics[width=\ww, height=\hh]{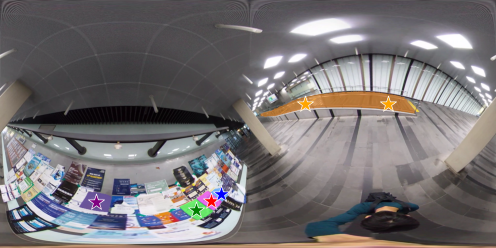}}
    \hfill
    \subfloat{\includegraphics[width=\ww, height=\hh]{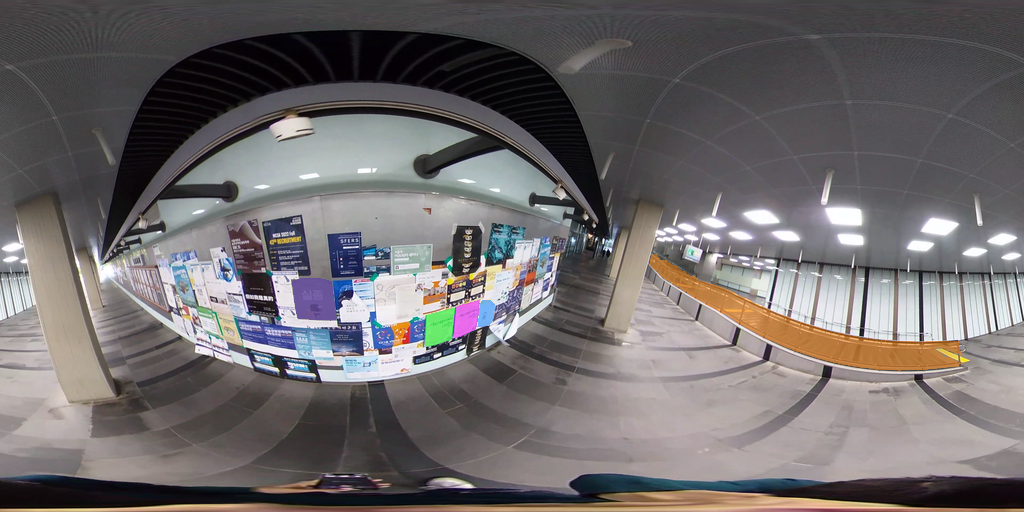}}
    \vspace{0.5mm}

    \addtocounter{subfigure}{-4}
    \subfloat[DEVA / Source view]{\includegraphics[width=\ww, height=\hh]{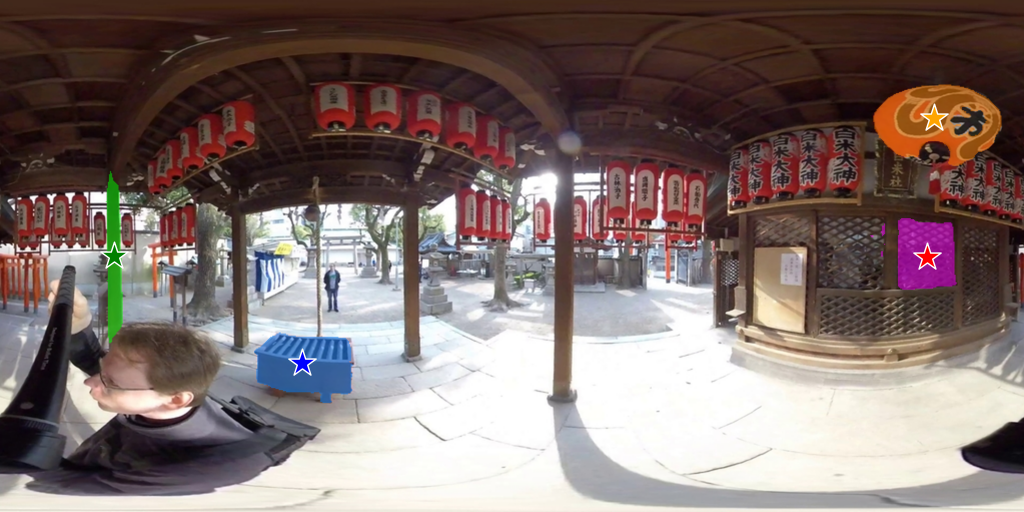}}
    \hfill
    \subfloat[DEVA / Target view]{\includegraphics[width=\ww, height=\hh]{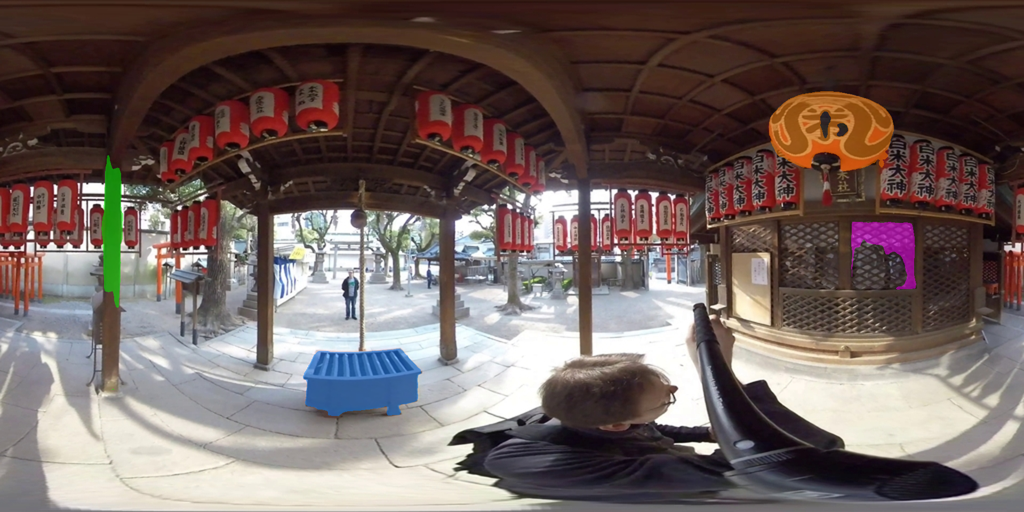}}
    \hfill
    \subfloat[OmniSplat / Source view]{\includegraphics[width=\ww, height=\hh]{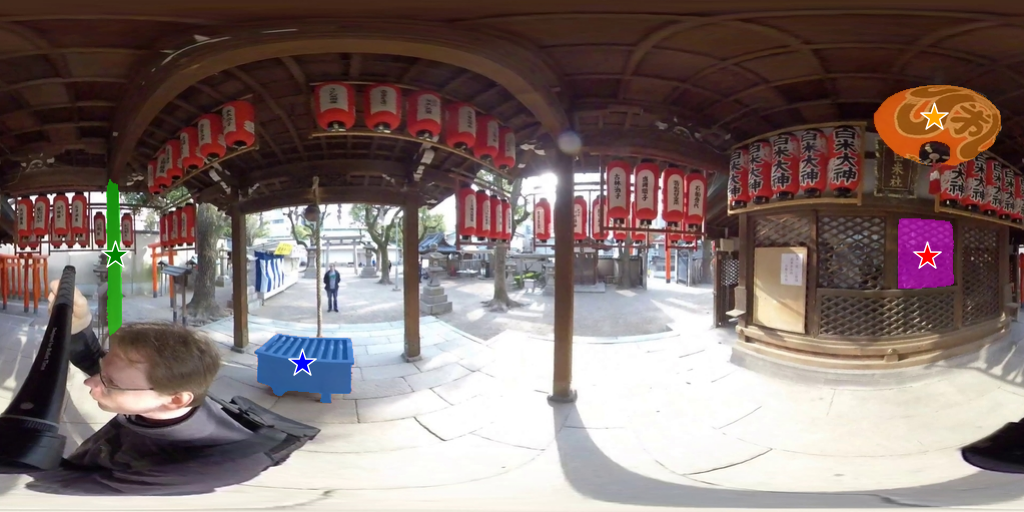}}
    \hfill
    \subfloat[OmniSplat / Target view]{\includegraphics[width=\ww, height=\hh]{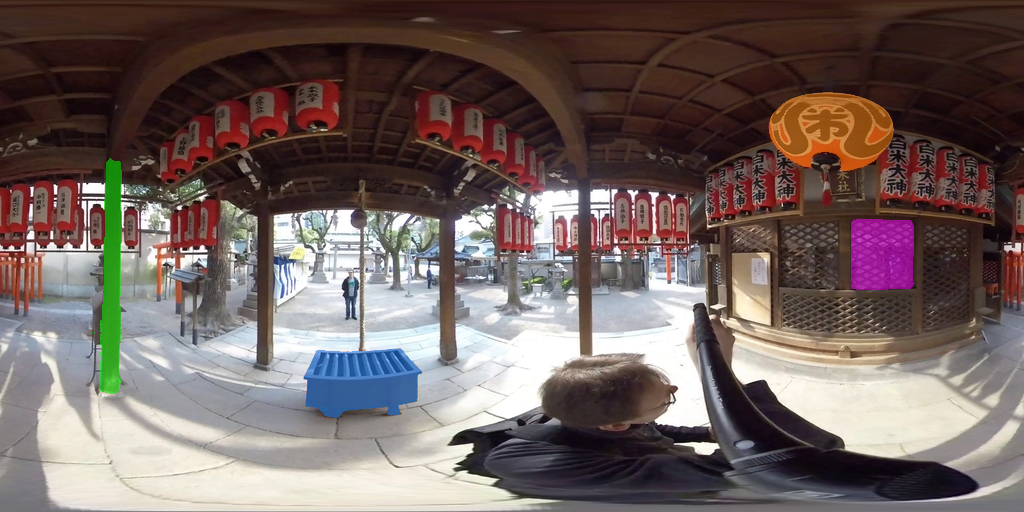}}
    \vspace{-1mm}
    \caption{
        \textbf{Visualization of segmentation matching.} We visualize the matched segments among the source and the target views.
        The stars in the image indicate query points for the user to segment objects containing the stars.
    }
    \vspace{-2mm}
    \label{fig:qual_editing_supple}
\end{figure*}
\begin{figure}[t]
    \newcommand{\ww}{0.495\linewidth}
    \newcommand{\hh}{0.333\linewidth}
    \centering
    \addtocounter{subfigure}{-2}
    \subfloat[Optimization-based Gaussians]{\includegraphics[width=\ww]{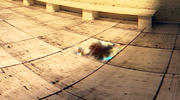}\label{fig:qual_editing_remove_a}}
    \hfill
    \subfloat[Pixel-aligned Gaussians]{\includegraphics[width=\ww]{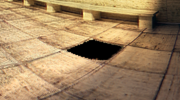}\label{fig:qual_editing_remove_b}}
    \vspace{-0.5mm}

    \caption{
        \textbf{Example of Gaussian removal results in the object segment.}
        Compared to (a), where overlapping Gaussians result in incomplete removal, (b) pixel-aligned Gaussians offer clear edges that are advantageous for editing.
    }
    \label{fig:qual_editing_remove}
    \vspace{-1mm}
\end{figure}

\section{Additional Gaussian Segmentation and Editing Results}

To facilitate 3D editing in the result Gaussians, we first construct multiview consistent segmentation maps using the cross-attention scores.
\Cref{fig:qual_editing_supple} presents additional examples of multiview consistent segmentation.
The selection operation allows users to choose regions by selecting Gaussians with the same label. 
Unlike optimization-based methods, pixel-aligned Gaussians provide cleaner boundary separation.
\Cref{fig:qual_editing_remove} illustrates the result of removing 3D Gaussians selected based on the multiview segmentation regions.
The removal result with the Gaussians optimized as 3DGS is illustrated in \Cref{fig:qual_editing_remove_a}.
Since the optimized Gaussians tend to have elongated shapes~\cite{xie2024physgaussian, hyung2024effective, edit1}, the selected regions are not cleanly removed, often resulting in needle-like artifacts. 
In contrast, our method with the pixel-aligned Gaussians provides clear boundaries, enabling a cleaner removal operation and establishing a solid foundation for subsequent operations such as image inpainting.

\section{Implementation Details and Additional Quantitative Results}

We write the indices for reference and test views for each scene in \Cref{tab:omniblender_performance_comparison,tab:ricoh_performance_comparison,tab:omniphotos_performance_comparison,tab:360roam_performance_comparison,tab:omniscenes_performance_comparison,tab:360vo_performance_comparison} along with detailed quantitative results.
For three datasets with relatively small camera motions (OmniBlender, Ricoh360~\cite{choi2023balanced}, OmniPhotos~\cite{bertel2020omniphotos}), we select the two reference views with large pose distances and set the rest as test views.
For the other three datasets (360Roam~\cite{huang2022tc360roam}, OmniScenes~\cite{kim2021piccolo}, 360VO~\cite{huang2022360vo}), the pose distance between two adjacent cameras is large, and we selected two frames with 3 to 4 timestep intervals as two reference views.
Then, the test view indices are composed of frames between the reference views.
For example, if the reference view indices are 1 and 5, the test view indices are 2, 3, and 4.

Also, we write quantitative metrics for every scene in \Cref{tab:omniblender_performance_comparison,tab:ricoh_performance_comparison,tab:omniphotos_performance_comparison,tab:360roam_performance_comparison,tab:omniscenes_performance_comparison,tab:360vo_performance_comparison}, respectively.
The best result for each metric is written in \textbf{bold}, and the second best result is written in \underline{underlined}.
As shown in the tables, OmniSplat+\textit{opt} shows the best or the second best results in almost all metrics.
Although ODGS shows conspicuous results, ODGS still shows optimal quality-runtime tradeoff, considering that ODGS requires scene-wise optimization.

\section{Additional Qualitative Results}

\begin{figure*}[ht]
    \newcommand{\ww}{0.198\linewidth}
    \centering
    \subfloat[Ground truth]{\includegraphics[width=\ww]{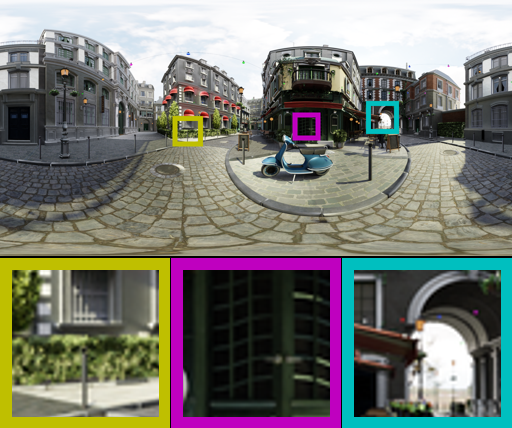}}
    \hfill
    \subfloat[PixelSplat (P)]{\includegraphics[width=\ww]{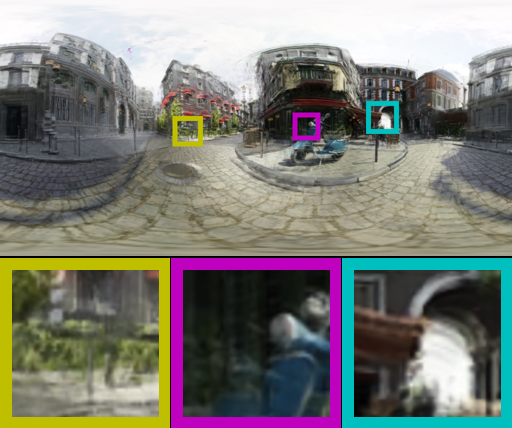}}
    \hfill
    \subfloat[MVSplat (P)]{\includegraphics[width=\ww]{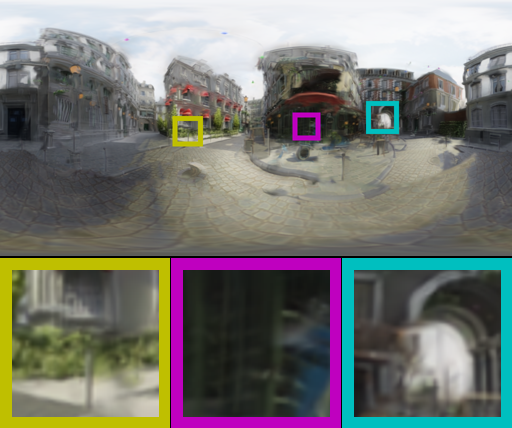}}
    \hfill
    \subfloat[MVSplat (Y)]{\includegraphics[width=\ww]{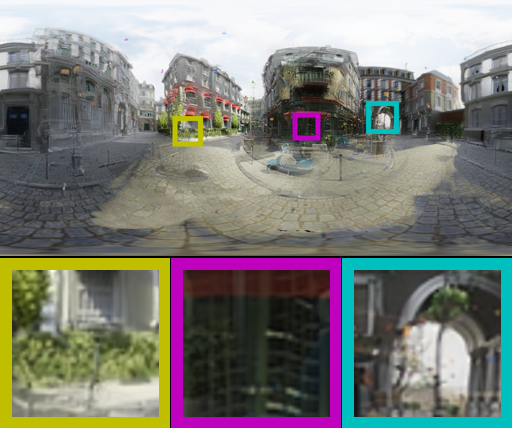}}
    \hfill
    \subfloat[OmniSplat]{\includegraphics[width=\ww]{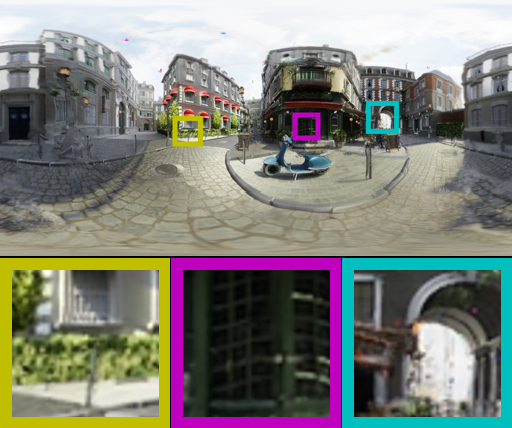}}
    
    \subfloat[ODGS]{\includegraphics[width=\ww]{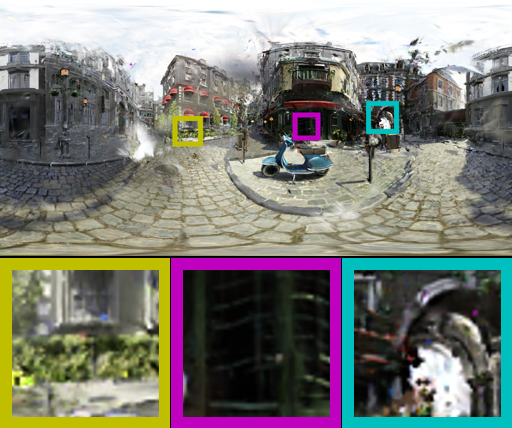}}
    \hfill
    \subfloat[PixelSplat (O)]{\includegraphics[width=\ww]{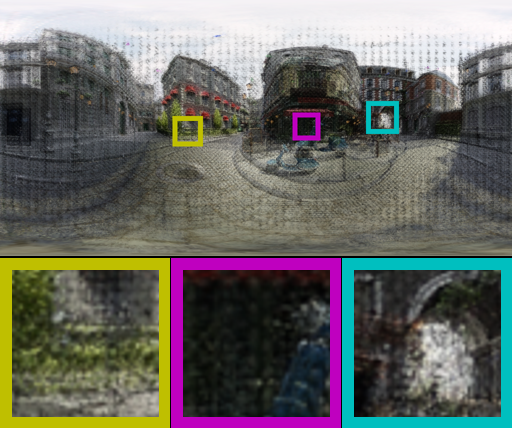}}
    \hfill
    \subfloat[MVSplat (O)]{\includegraphics[width=\ww]{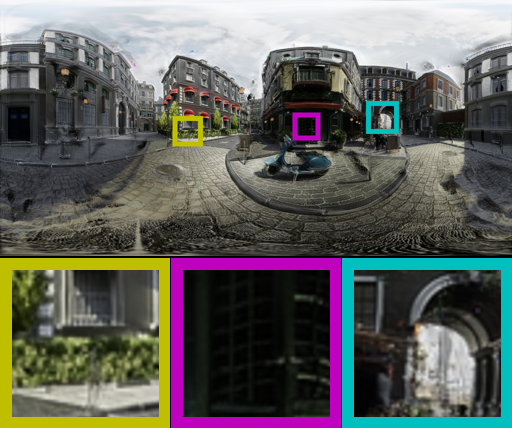}}
    \hfill
    \subfloat[DepthSplat (O)]{\includegraphics[width=\ww]{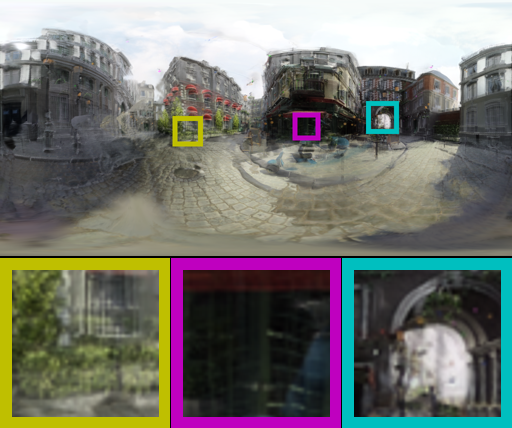}}
    \hfill
    \subfloat[OmniSplat + \textit{opt}]{\includegraphics[width=\ww]{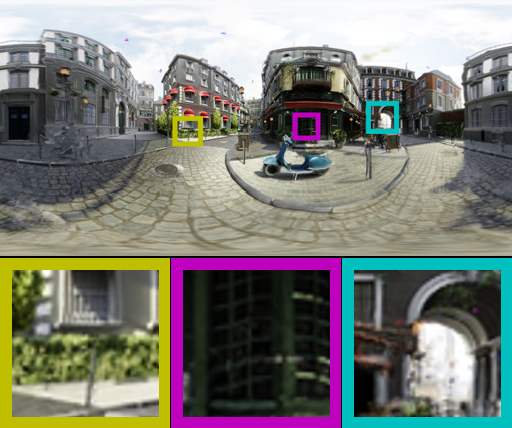}}
    \vspace{4mm}

    \caption{
        Qualitative comparison on OmniBlender dataset.
    }
    \label{fig:supp_omniblender}
\end{figure*}

\begin{figure*}[ht]
    \newcommand{\ww}{0.198\linewidth}
    \centering
    \subfloat[Ground truth]{\includegraphics[width=\ww]{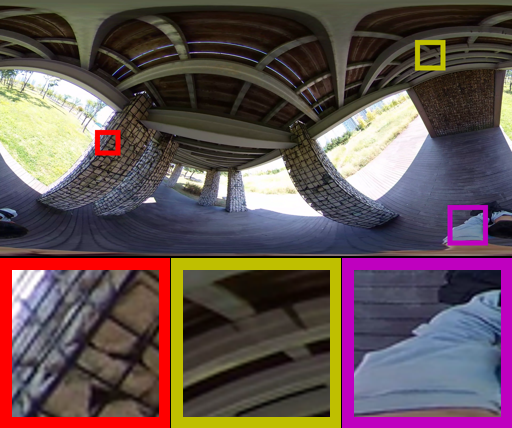}}
    \hfill
    \subfloat[PixelSplat (P)]{\includegraphics[width=\ww]{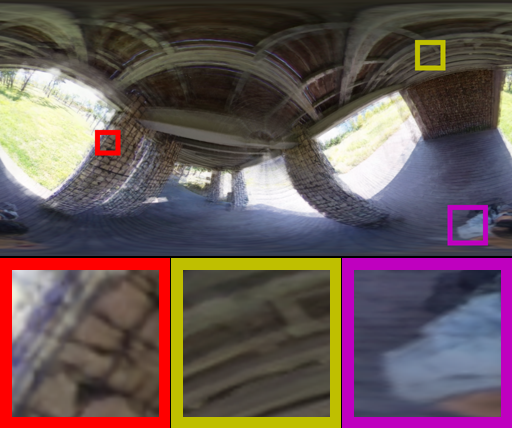}}
    \hfill
    \subfloat[MVSplat (P)]{\includegraphics[width=\ww]{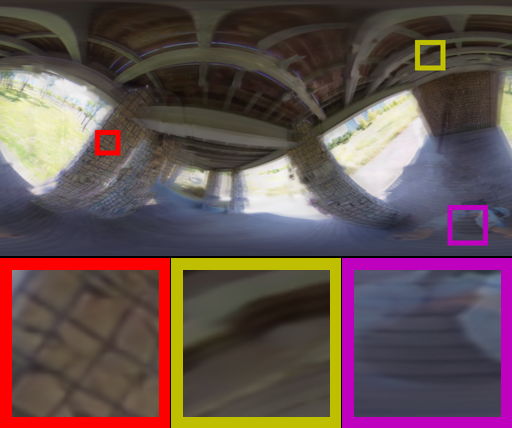}}
    \hfill
    \subfloat[MVSplat (Y)]{\includegraphics[width=\ww]{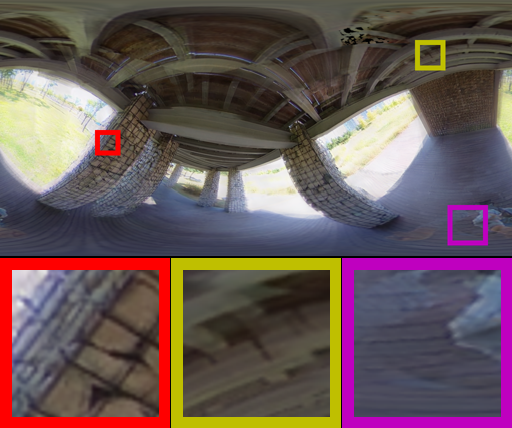}}
    \hfill
    \subfloat[OmniSplat]{\includegraphics[width=\ww]{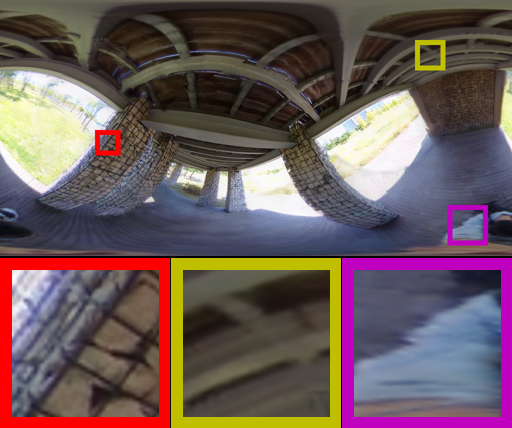}}
    
    \subfloat[ODGS]{\includegraphics[width=\ww]{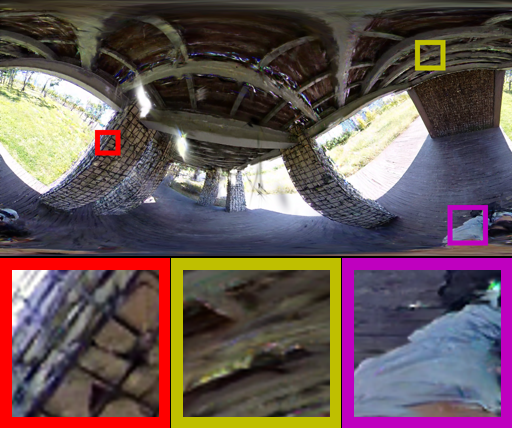}}
    \hfill
    \subfloat[PixelSplat (O)]{\includegraphics[width=\ww]{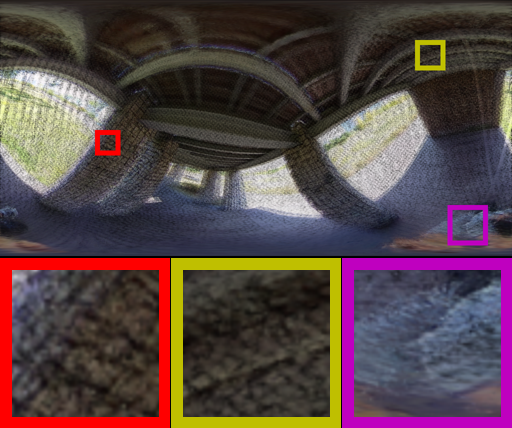}}
    \hfill
    \subfloat[MVSplat (O)]{\includegraphics[width=\ww]{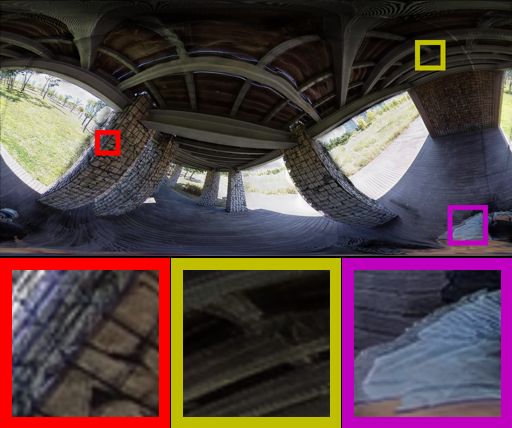}}
    \hfill
    \subfloat[DepthSplat (O)]{\includegraphics[width=\ww]{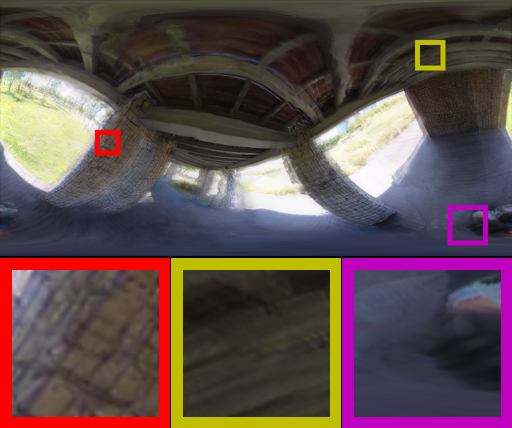}}
    \hfill
    \subfloat[OmniSplat + \textit{opt}]{\includegraphics[width=\ww]{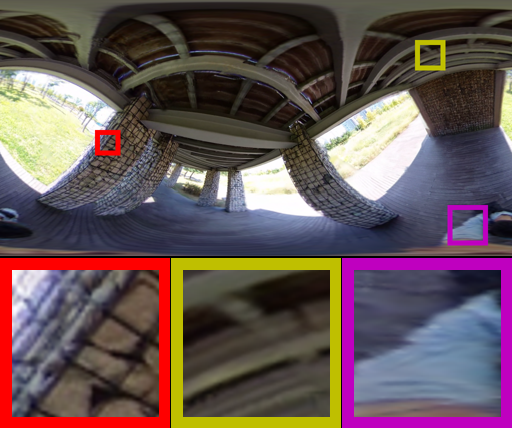}}
    \vspace{4mm}

    \caption{
        Qualitative comparison on Ricoh dataset.
    }
    \label{fig:supp_ricoh}
\end{figure*}

\begin{figure*}[ht]
    \newcommand{\ww}{0.198\linewidth}
    \centering
    \subfloat[Ground truth]{\includegraphics[width=\ww]{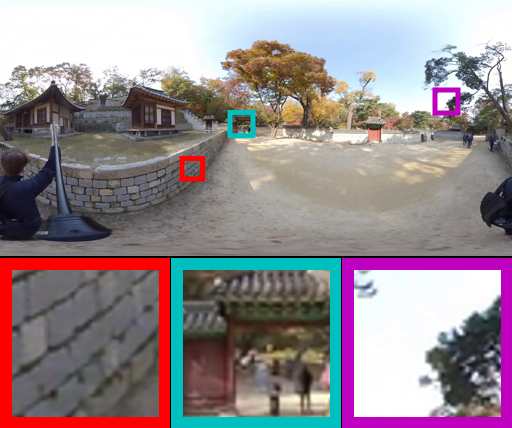}}
    \hfill
    \subfloat[PixelSplat (P)]{\includegraphics[width=\ww]{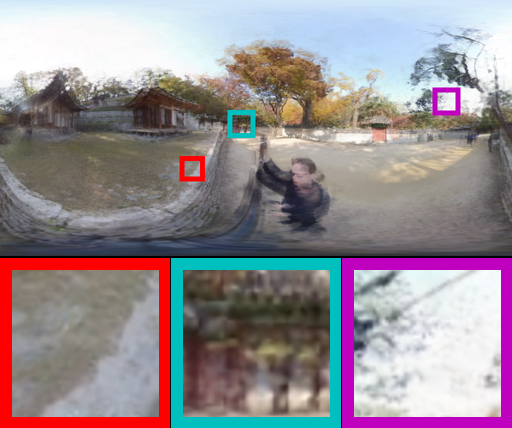}}
    \hfill
    \subfloat[MVSplat (P)]{\includegraphics[width=\ww]{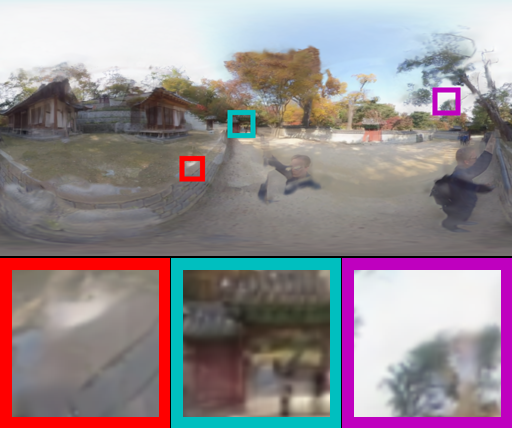}}
    \hfill
    \subfloat[MVSplat (Y)]{\includegraphics[width=\ww]{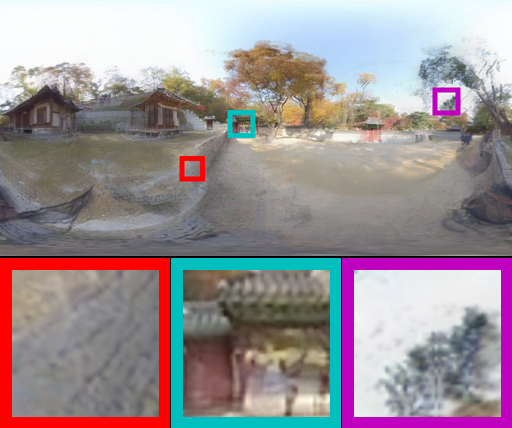}}
    \hfill
    \subfloat[OmniSplat]{\includegraphics[width=\ww]{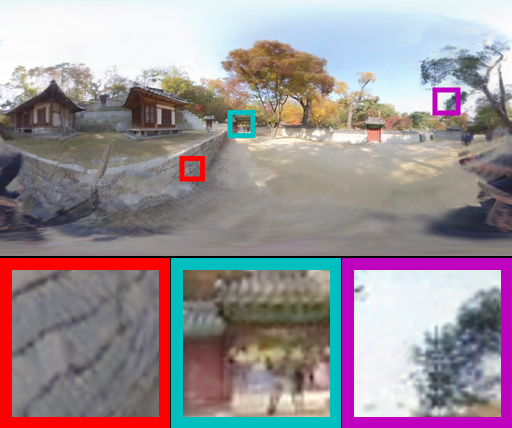}}
    
    \subfloat[ODGS]{\includegraphics[width=\ww]{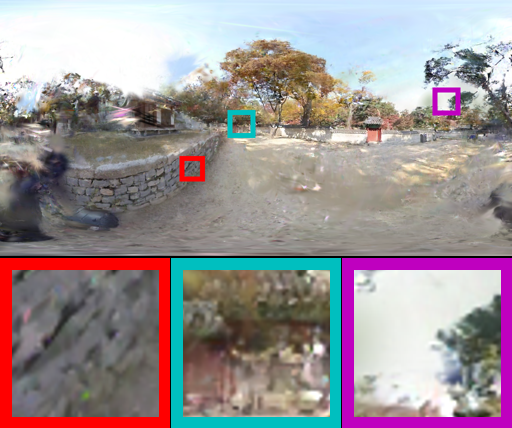}}
    \hfill
    \subfloat[PixelSplat (O)]{\includegraphics[width=\ww]{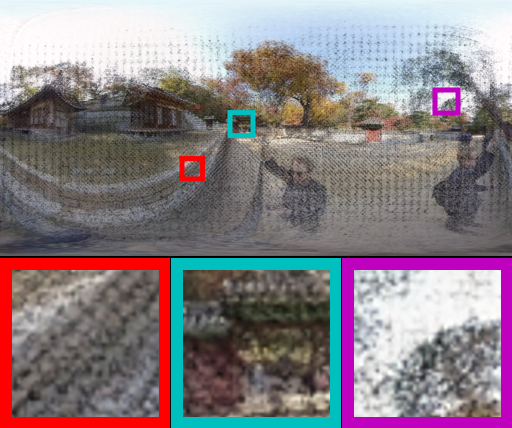}}
    \hfill
    \subfloat[MVSplat (O)]{\includegraphics[width=\ww]{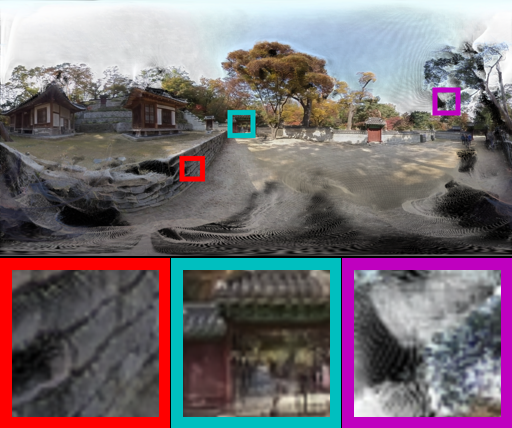}}
    \hfill
    \subfloat[DepthSplat (O)]{\includegraphics[width=\ww]{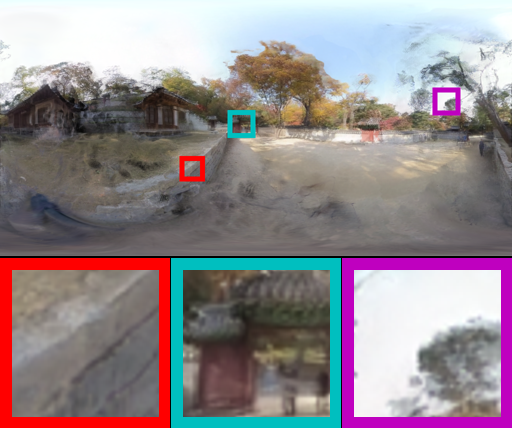}}
    \hfill
    \subfloat[OmniSplat + \textit{opt}]{\includegraphics[width=\ww]{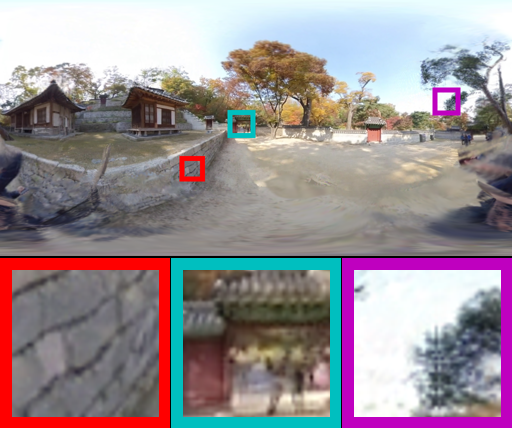}}
    \vspace{4mm}

    \caption{
        Qualitative comparison on OmniPhotos dataset.
    }
    \label{fig:supp_omniphotos}
\end{figure*}

\begin{figure*}[ht]
    \newcommand{\ww}{0.198\linewidth}
    \centering
    
    \subfloat[Ground truth]{\includegraphics[width=\ww]{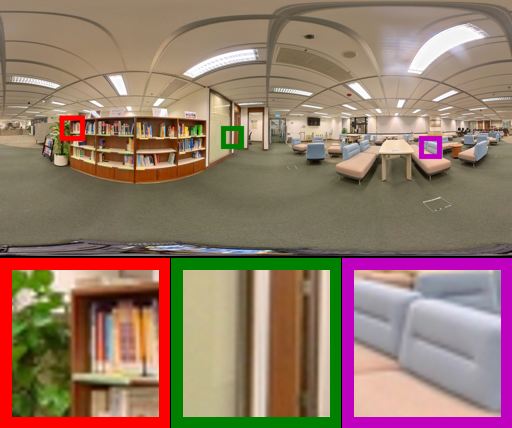}}
    \hfill
    \subfloat[PixelSplat (P)]{\includegraphics[width=\ww]{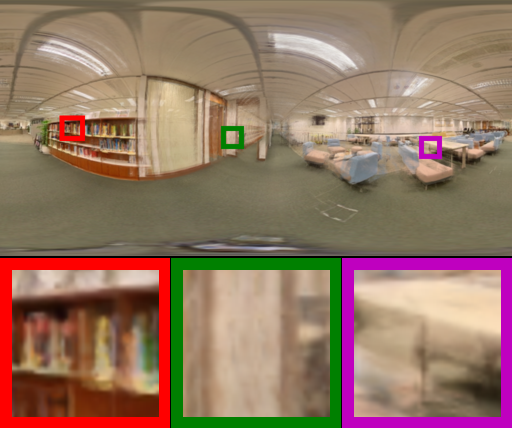}}
    \hfill
    \subfloat[MVSplat (P)]{\includegraphics[width=\ww]{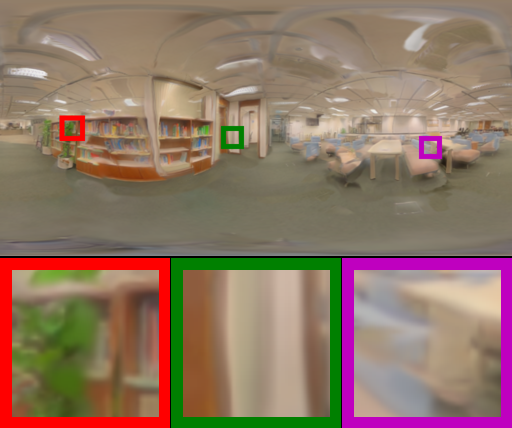}}
    \hfill
    \subfloat[MVSplat (Y)]{\includegraphics[width=\ww]{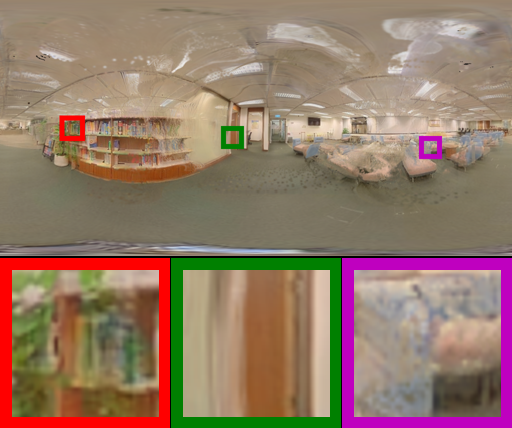}}
    \hfill
    \subfloat[OmniSplat]{\includegraphics[width=\ww]{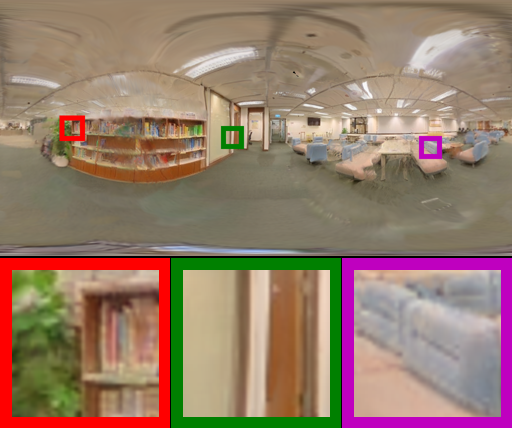}}
    
    \subfloat[ODGS]{\includegraphics[width=\ww]{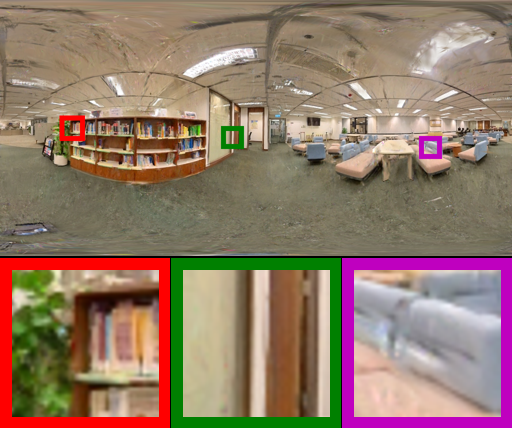}}
    \hfill
    \subfloat[PixelSplat (O)]{\includegraphics[width=\ww]{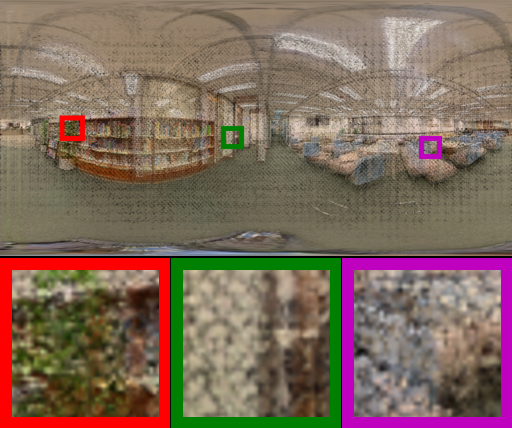}}
    \hfill
    \subfloat[MVSplat (O)]{\includegraphics[width=\ww]{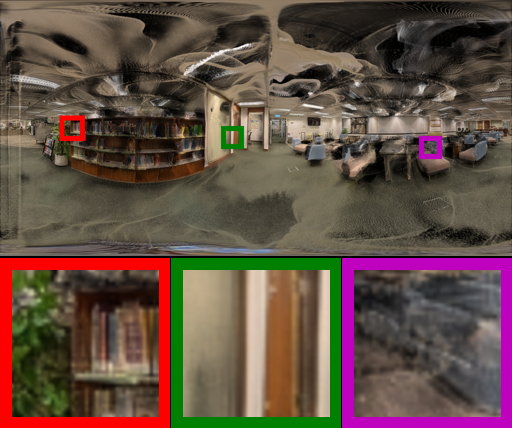}}
    \hfill
    \subfloat[DepthSplat (O)]{\includegraphics[width=\ww]{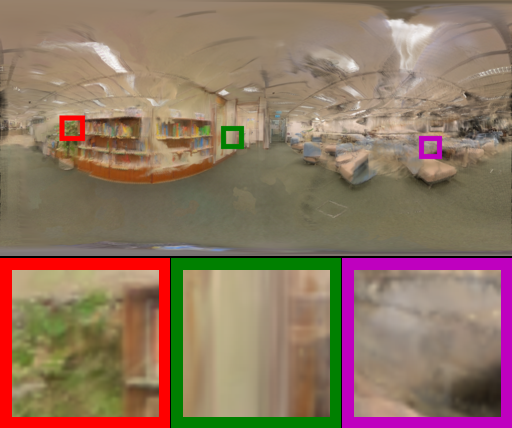}}
    \hfill
    \subfloat[OmniSplat + \textit{opt}]{\includegraphics[width=\ww]{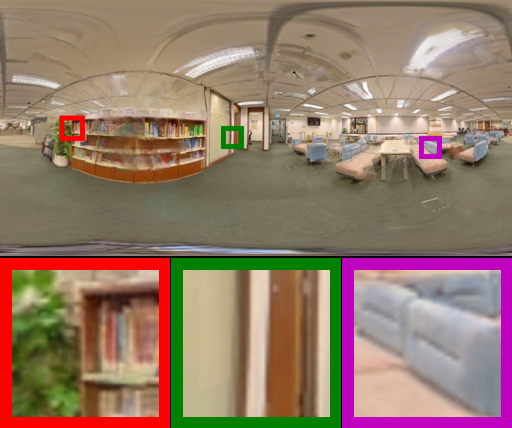}}

    \caption{
        Qualitative comparison on 360Roam dataset.
    }
    \label{fig:supp_360roam}
\end{figure*}

\begin{figure*}[ht]
    \newcommand{\ww}{0.198\linewidth}
    \centering
    \subfloat[Ground truth]{\includegraphics[width=\ww]{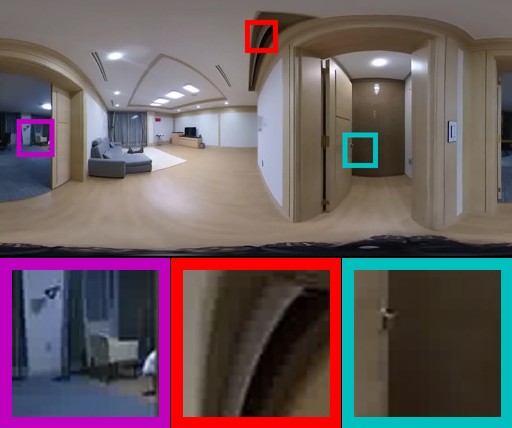}}
    \hfill
    \subfloat[PixelSplat (P)]{\includegraphics[width=\ww]{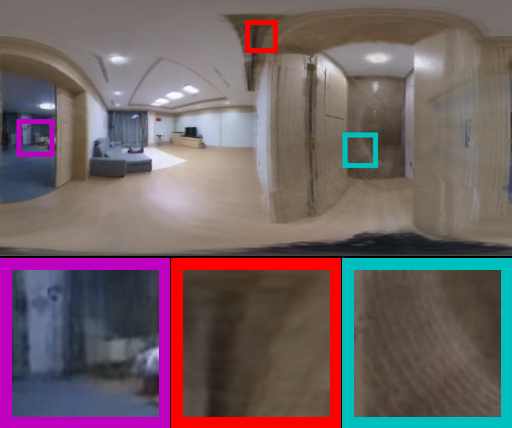}}
    \hfill
    \subfloat[MVSplat (P)]{\includegraphics[width=\ww]{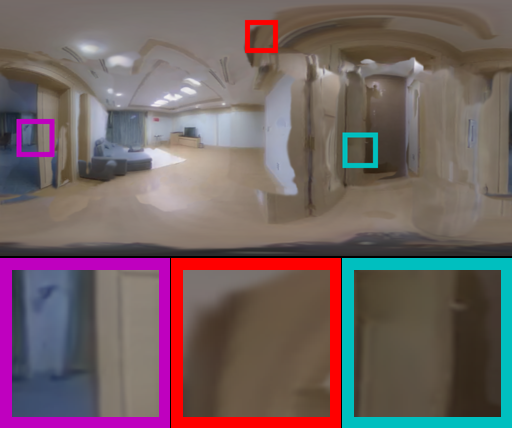}}
    \hfill
    \subfloat[MVSplat (Y)]{\includegraphics[width=\ww]{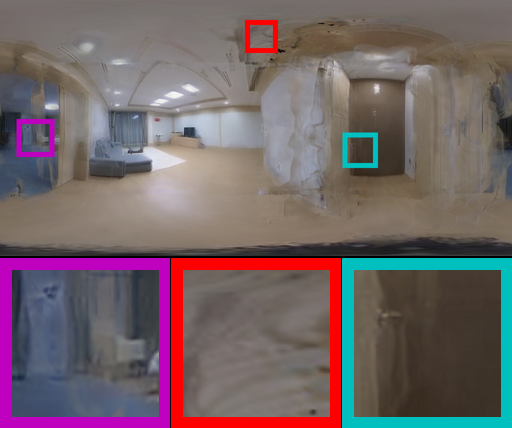}}
    \hfill
    \subfloat[OmniSplat]{\includegraphics[width=\ww]{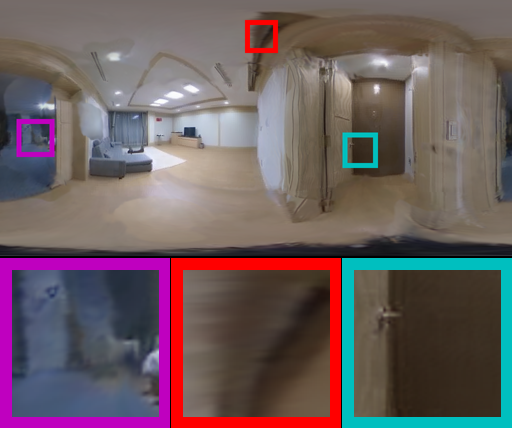}}
    
    \subfloat[ODGS]{\includegraphics[width=\ww]{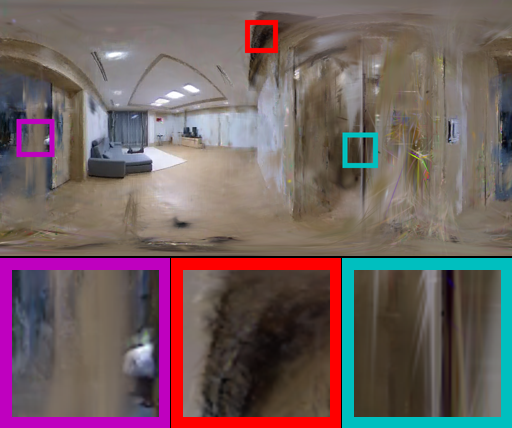}}
    \hfill
    \subfloat[PixelSplat (O)]{\includegraphics[width=\ww]{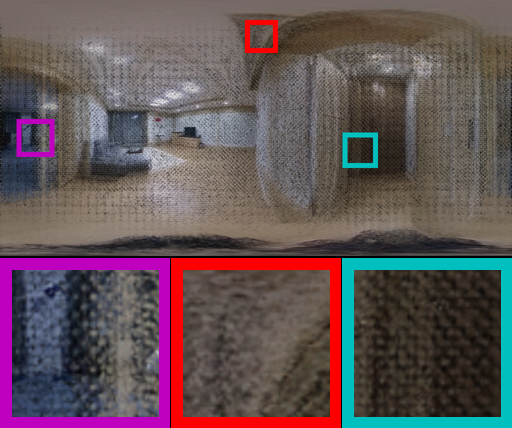}}
    \hfill
    \subfloat[MVSplat (O)]{\includegraphics[width=\ww]{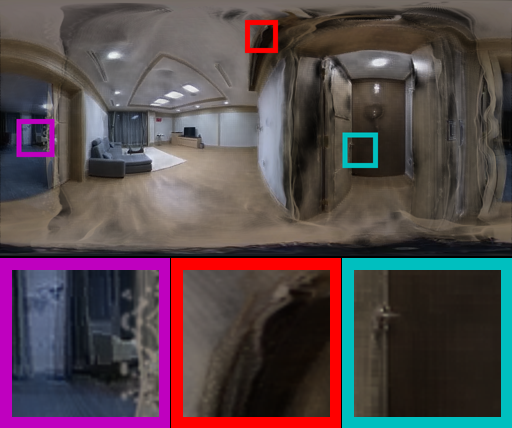}}
    \hfill
    \subfloat[DepthSplat (O)]{\includegraphics[width=\ww]{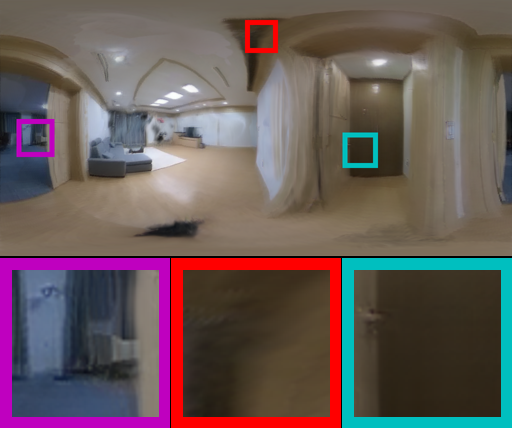}}
    \hfill
    \subfloat[OmniSplat + \textit{opt}]{\includegraphics[width=\ww]{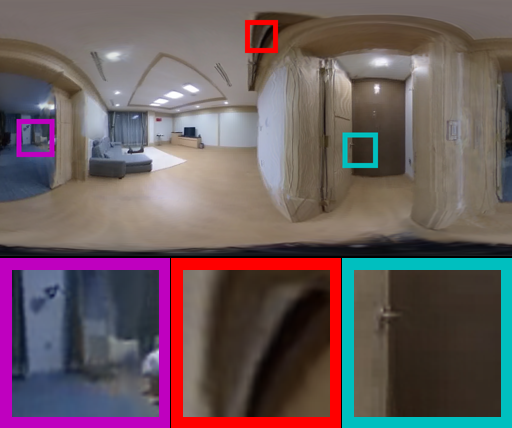}}

    \caption{
        Qualitative comparison on OmniScenes dataset.
    }
    \label{fig:supp_omniscenes}
\end{figure*}

\begin{figure*}[ht]
    \newcommand{\ww}{0.198\linewidth}
    \centering
    \subfloat[Ground truth]{\includegraphics[width=\ww]{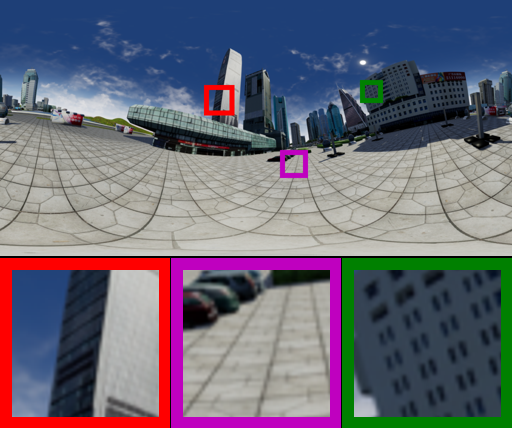}}
    \hfill
    \subfloat[PixelSplat (P)]{\includegraphics[width=\ww]{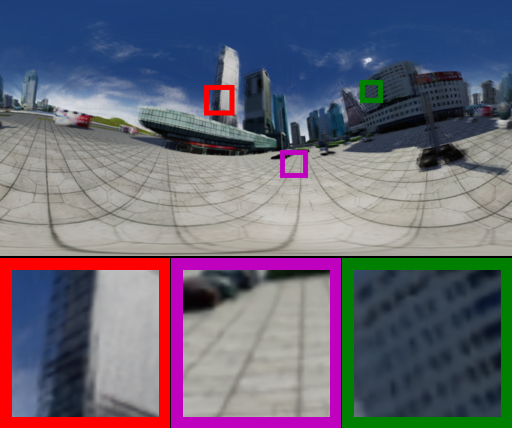}}
    \hfill
    \subfloat[MVSplat (P)]{\includegraphics[width=\ww]{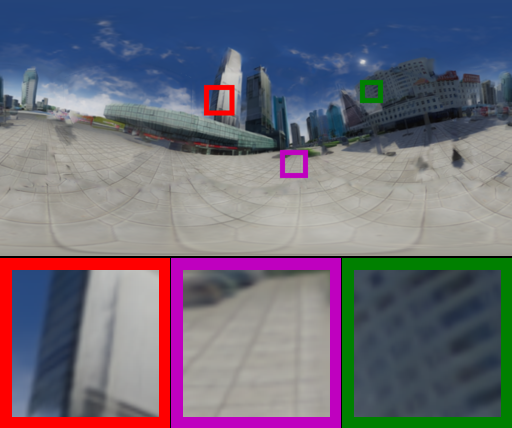}}
    \hfill
    \subfloat[MVSplat (Y)]{\includegraphics[width=\ww]{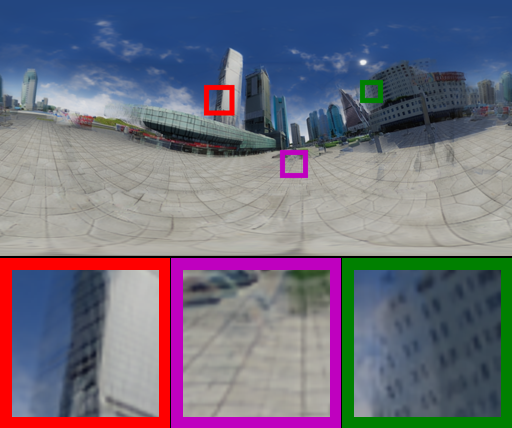}}
    \hfill
    \subfloat[OmniSplat]{\includegraphics[width=\ww]{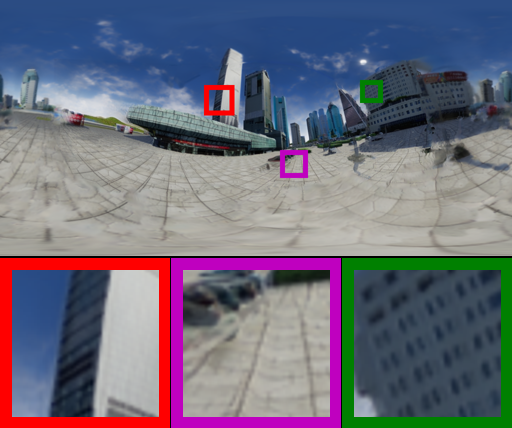}}
    
    \subfloat[ODGS]{\includegraphics[width=\ww]{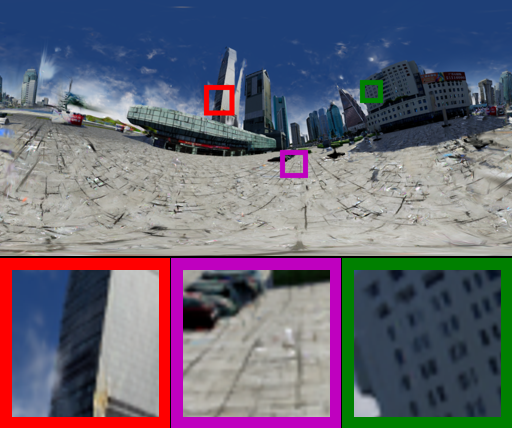}}
    \hfill
    \subfloat[PixelSplat (O)]{\includegraphics[width=\ww]{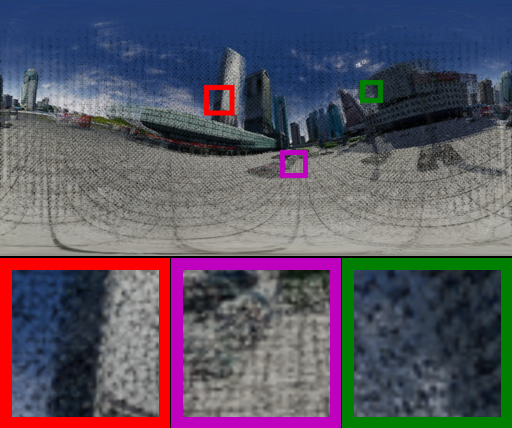}}
    \hfill
    \subfloat[MVSplat (O)]{\includegraphics[width=\ww]{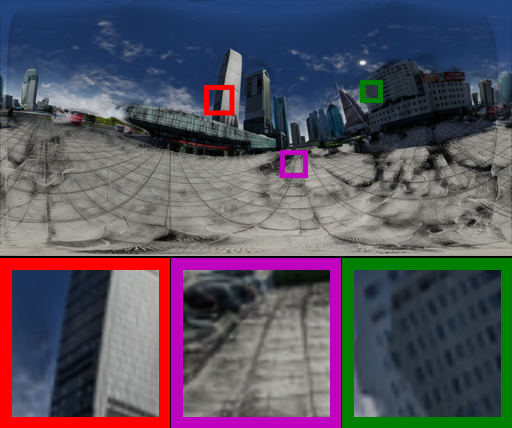}}
    \hfill
    \subfloat[DepthSplat (O)]{\includegraphics[width=\ww]{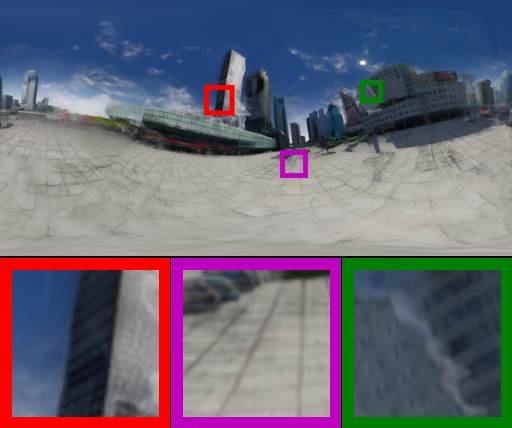}}
    \hfill
    \subfloat[OmniSplat + \textit{opt}]{\includegraphics[width=\ww]{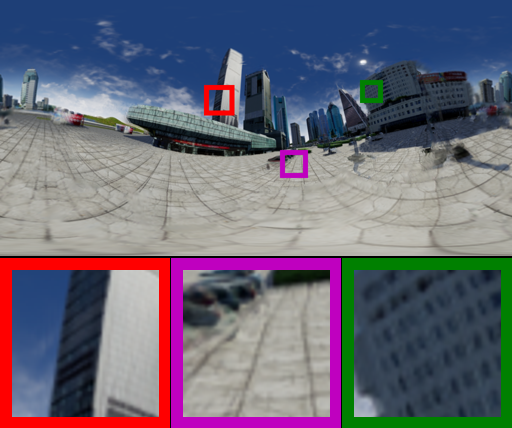}}

    \caption{
        Qualitative comparison on 360VO dataset.
    }
    \label{fig:supp_360vo}
\end{figure*}

In \Cref{fig:supp_omniblender,fig:supp_ricoh,fig:supp_omniphotos,fig:supp_360roam,fig:supp_omniscenes,fig:supp_360vo}, we provide additional qualitative comparison with more baselines on various datasets.
Perspective feed-forward models, PixelSplat~(P)~\cite{charatan2024pixelsplat} and MVSplat~(P)~\cite{chen2024mvsplat}, produce blurry images.
For PixelSplat, Gaussians are generated for every pair among twelve perspective images, and since the positions of the Gaussians generated for each pair are slightly different, a blurry image is rendered in the novel viewpoints.
For MVSplat, the encoded features and cross-attention values are averaged across all perspective images, which causes blur or ghost artifacts in novel view synthesized images.
Omnidirectional feed-forward models, PixelSplat (O), MVSplat (O), and DepthSplat (O), produce dark images overall and striped patterns in the upper and lower areas.
We attribute the phenomenon to the non-uniform sampling of the omnidirectional grid during rasterization.
ODGS sometimes shows prominent image quality, but it requires over 30 minutes of optimization for each scene.
Moreover, it often falls into overfitting, generating floater artifacts that severely degrade the quality of the image.
Our method, OmniSplat, produces clearer and more accurate images than other feed-forward networks and ODGS.
Also, the results can be further improved with a small number of optimizations (OmniSplat+\textit{opt}), which only takes 12 seconds, which is shorter than the execution time of PixelSplat (P).

\clearpage

\begin{table*}[h]
    \centering
    \resizebox{\textwidth}{!}{
    \begin{tabular}{c|c|c|c||c|c|c|c|c|c|c|c|c}
        \toprule[1.0pt]
        Scene & Ref. view & Test view & ODGS & PixelSplat (P) & LatentSplat (P) & MVSplat (P) & PixelSplat (O) & MVSplat (O) & DepthSplat (O) & MVSplat (Y) & OmniSplat & OmniSplat+\textit{opt} \\
        \midrule
        archiviz-flat & 14, 86 & 2, 6, ..., 98
            & 22.17 / 0.7511 / 0.3188
            & 20.93 / 0.7128 / 0.3274
            & 18.40 / 0.6859 / 0.3981
            & 18.35 / 0.6704 / 0.3870
            & 17.28 / 0.4285 / 0.6206
            & 17.73 / 0.6818 / 0.4138
            & 19.06 / 0.6812 / 0.4491
            & 19.76 / 0.7013 / 0.3740
            & \underline{23.93} / \underline{0.8010} / \underline{0.2688}
            & \textbf{24.97} / \textbf{0.8147} / \textbf{0.2618} \\
        barbershop & 14, 86 & 2, 6, ..., 98 
            & 25.01 / 0.7683 / 0.2802
            & 19.83 / 0.6243 / 0.3588
            & 17.96 / 0.5627 / 0.4666
            & 17.98 / 0.5636 / 0.4445 
            & 17.70 / 0.4262 / 0.6197 
            & 16.60 / 0.5214 / 0.4459 
            & 18.13 / 0.5510 / 0.5534
            & 18.33 / 0.5615 / 0.4885
            & \underline{21.76} / \underline{0.7067} / \underline{0.3328}
            & \textbf{22.70} / \textbf{0.7279} / \textbf{0.3110} \\
        bistro bike & 14, 86 & 2, 6, ..., 98 
            & 21.05 / 0.6800 / 0.2687
            & \underline{20.20} / 0.5882 / 0.2981
            & 15.43 / 0.4248 / 0.4343
            & 17.29 / 0.4619 / 0.4241 
            & 16.50 / 0.3613 / 0.5527 
            & 16.40 / 0.5253 / 0.3354 
            & 16.97 / 0.4106 / 0.4817
            & 17.54 / 0.4463 / 0.4007
            & \underline{20.20} / \underline{0.6312} / \underline{0.2844}
            & \textbf{21.75} / \underline{0.6868} / \textbf{0.2459} \\
        bistro square & 14, 86 & 2, 6, ..., 98 
            & 19.30 / 0.6397 / 0.3030
            & \textbf{20.11} / \textbf{0.5989} / \underline{0.2904}
            & 15.43 / 0.4290 / 0.4427
            & 16.41 / 0.4348 / 0.4376
            & 14.93 / 0.2878 / 0.5746 
            & 13.76 / 0.4338 / 0.4070 
            & 15.36 / 0.3853 / 0.5050
            & 16.11 / 0.4381 / 0.4085
            & 18.43 / 0.5535 / 0.3185 
            & \underline{18.98} / \underline{0.5845} / \textbf{0.2844} \\
        classroom & 14, 86 & 2, 6, ..., 98 
            & 23.55 / 0.6436 / 0.3655
            & 20.50 / 0.6803 / \underline{0.3112}
            & 17.46 / 0.6150 / 0.4574
            & 16.47 / 0.6022 / 0.4719
            & 16.61 / 0.3884 / 0.6115 
            & 16.30 / 0.5223 / 0.4130 
            & 17.23 / 0.5620 / 0.5403
            & 17.97 / 0.6018 / 0.4537
            & \underline{20.91} / \underline{0.7113} / 0.3505 
            & \textbf{21.63} / \textbf{0.7254} / \textbf{0.3107} \\
        fisher hut & 14, 86 & 2, 6, ..., 98 
            & 21.07 / 0.6326 / 0.3516
            & 22.75 / \textbf{0.7389} / 0.3253
            & 19.91 / 0.6673 / 0.3773
            & 21.08 / 0.7226 / 0.3660
            & 18.89 / 0.4390 / 0.5900 
            & 20.90 / 0.6487 / 0.4265 
            & 21.83 / 0.6690 / 0.4156
            & 21.79 / 0.6726 / 0.3644
            & \underline{25.08} / 0.7244 / \underline{0.3155}
            & \textbf{26.60} / \underline{0.7269} / \textbf{0.2743} \\
        lone monk & 14, 86 & 2, 6, ..., 98 
            & 20.21 / 0.6907 / 0.3053
            & 16.89 / 0.5907 / 0.3402
            & 14.12 / 0.4640 / 0.4277
            & 15.33 / 0.4947 / 0.4195
            & 14.23 / 0.3075 / 0.6029 
            & 13.09 / 0.5371 / 0.4259 
            & 15.69 / 0.4690 / 0.4417
            & 16.51 / 0.5120 / 0.3855
            & \underline{19.04} / \underline{0.6175} / \underline{0.2978} 
            & \textbf{19.28} / \textbf{0.6374} / \textbf{0.2945} \\
        LOU & 14, 86 & 2, 6, ..., 98 
            & 21.53 / 0.7332 / 0.2469
            & 18.97 / \underline{0.6801} / 0.3246
            & 14.47 / 0.5251 / 0.4452
            & 15.17 / 0.4348 / 0.4459 
            & 16.49 / 0.4040 / 0.5842
            & 18.16 / 0.6678 / 0.3141
            & 16.53 / 0.5327 / 0.4082
            & 16.72 / 0.4991 / 0.3821
            & \underline{19.51} / 0.6438 / \underline{0.3071}
            & \textbf{20.33} / \textbf{0.7420} / \textbf{0.2622} \\
        pavilion midday chair & 14, 86 & 2, 6, ..., 98 
            & 21.24 / 0.6541 / 0.3516
            & \underline{21.28} / \textbf{0.7035} / \underline{0.3198}
            & 18.29 / 0.6050 / 0.4178
            & 18.30 / 0.6470 / 0.3868
            & 17.06 / 0.4382 / 0.5765
            & 16.71 / 0.5623 / 0.4193
            & 17.86 / 0.6136 / 0.4251
            & 18.63 / 0.6294 / 0.3887
            & 21.02 / 0.6836 / 0.3216
            & \textbf{22.09} / \underline{0.7005} / \textbf{0.3051} \\
        pavilion midday pond & 14, 86 & 2, 6, ..., 98 
            & 19.21 / 0.4918 / 0.4229
            & \textbf{19.70} / \textbf{0.6058} / \underline{0.3353}
            & 15.31 / 0.4486 / 0.5241
            & 14.24 / 0.4348 / 0.6142
            & 15.40 / 0.3021 / 0.6316
            & 15.63 / 0.4752 / 0.4120
            & 16.46 / 0.4528 / 0.4974
            & 16.24 / 0.4902 / 0.4407
            & 18.34 / 0.5609 / 0.3776
            & \underline{19.18} / \underline{0.5749} / \textbf{0.3328} \\
        restroom & 14, 86 & 2, 6, ..., 98 
            & 30.21 / 0.8030 / 0.2220
            & \underline{24.95} / 0.6947 / \underline{0.3132}
            & 23.33 / 0.6481 / 0.4063
            & 21.71 / 0.6551 / 0.4263 
            & 21.54 / 0.4381 / 0.5999
            & 21.43 / 0.6229 / 0.3305
            & 21.67 / 0.5730 / 0.5103
            & 23.16 / 0.6379 / 0.4304
            & 23.02 / \underline{0.7266} / 0.3792
            & \textbf{28.17} / \textbf{0.7720} / \textbf{0.2845} \\
        \midrule
        \midrule
        average & - & - 
            & 22.23 / 0.6807 / 0.3124
            & 20.56 / 0.6562 / \underline{0.3222}
            & 17.28 / 0.5523 / 0.4361
            & 17.48 / 0.5565 / 0.4385 
            & 16.97 / 0.3837 / 0.5967
            & 16.97 / 0.5635 / 0.3949
            & 17.89 / 0.5364 / 0.4753
            & 18.43 / 0.5627 / 0.4107
            & \underline{21.02} / \underline{0.6691} / 0.3231
            & \textbf{22.33} / \textbf{0.6994} / \textbf{0.2879} \\
        \bottomrule
    \end{tabular}
    }
    \caption{Scene-wise quantitative results of 3D reconstruction on \textbf{OmniBlender} dataset.
    }
    \label{tab:omniblender_performance_comparison}
\end{table*}

\begin{table*}[h]
    \centering
    \resizebox{\textwidth}{!}{
    \begin{tabular}{c|c|c|c||c|c|c|c|c|c|c|c|c}
        \toprule[1.0pt]
        Scene & Ref. view & Test view & ODGS & PixelSplat (P) & LatentSplat (P) & MVSplat (P) & PixelSplat (O) & MVSplat (O) & DepthSplat (O) & MVSplat (Y) & OmniSplat & OmniSplat+\textit{opt} \\
        \midrule
        bricks & 35, 69 & 1, 3, ..., 99
            & 16.41 / 0.5116 / 0.3738
            & 16.28 / 0.5161 / 0.4249
            & 13.38 / 0.4586 / 0.4781
            & 17.86 / 0.5834 / 0.3820
            & 15.89 / 0.4091 / 0.5296
            & 15.01 / 0.5017 / 0.4070
            & 13.34 / 0.3861 / 0.5950
            & 16.76 / 0.5072 / 0.4053
            & \underline{19.25} / \underline{0.5980} / \underline{0.3374}
            & \textbf{19.91} / \textbf{0.6388} / \textbf{0.3052}  \\
        bridge & 33, 57 & 1, 3, ..., 99
            & 15.84 / 0.4353 / 0.4546
            & 16.45 / 0.5264 / 0.3999
            & 15.71 / 0.4367 / 0.5402
            & 18.68 / \underline{0.6120} / 0.3446
            & 16.62 / 0.3937 / 0.5269
            & 13.83 / 0.4450 / 0.4677
            & 13.41 / 0.4371 / 0.5585
            & 16.90 / 0.5111 / 0.3976
            & \underline{19.13} / 0.5947 / \underline{0.3436}
            & \textbf{19.53} / \textbf{0.6198} / \textbf{0.3183}  \\
        bridge under & 23, 77 & 1, 3, ..., 99
            & 18.57 / 0.4841 / 0.3916
            & 15.78 / 0.4519 / 0.4572
            & 14.67 / 0.4772 / 0.4473
            & 14.24 / 0.4492 / 0.4896
            & 15.51 / 0.3123 / 0.5839
            & 16.08 / 0.4179 / 0.4140
            & 16.72 / 0.4394 / 0.4879
            & 17.57 / 0.4607 / 0.4183
            & \underline{18.95} / \underline{0.5598} / \underline{0.3572}
            & \textbf{19.52} / \textbf{0.5794} / \textbf{0.3279} \\
        cat tower & 3, 83 & 1, 3, ..., 99
            & 15.09 / 0.4641 / 0.4210
            & \textbf{20.19} / \textbf{0.7243} / \textbf{0.3186}
            & 17.14 / 0.4167 / 0.6799
            & 17.25 / \underline{0.6241} / 0.3766
            & 16.68 / 0.3654 / 0.5489
            & 16.53 / 0.4640 / 0.4855
            & 14.83 / 0.4511 / 0.5021
            & 16.05 / 0.4910 / 0.4573
            & 18.65 / 0.5348 / 0.4005
            & \underline{19.42} / 0.5507 / \underline{0.3592}  \\
        center & 25, 49 & 1, 3, ..., 99
            & 19.56 / 0.6045 / 0.3659
            & \textbf{21.61} / 0.6936 / \underline{0.2948}
            & 17.28 / 0.3716 / 0.5863
            & 21.32 / \textbf{0.7594} / 0.3107
            & 17.97 / 0.4407 / 0.5556  
            & 17.01 / 0.6090 / 0.3886  
            & 16.38 / 0.5985 / 0.4985
            & 19.15 / 0.6708 / 0.3459
            & 20.66 / 0.7094 / 0.3210  
            & \underline{21.57} / \underline{0.7320} / \textbf{0.2887}  \\
        farm & 83, 99 & 1, 3, ..., 99
            & 18.05 / 0.5087 / 0.3431
            & 18.31 / 0.5168 / 0.3605
            & 16.31 / 0.4110 / 0.4937
            & \textbf{19.73} / \textbf{0.5789} / \underline{0.3447}
            & 16.49 / 0.3470 / 0.5068
            & 12.93 / 0.3631 / 0.4969
            & 17.22 / 0.4613 / 0.4171
            & 17.76 / 0.4940 / 0.3723
            & 18.29 / 0.5133 / 0.3646
            & \underline{18.56} / \underline{0.5227} / \textbf{0.3409}  \\
        flower & 29, 55 & 1, 3, ..., 99
            & 15.37 / 0.4163 / 0.4084
            & \underline{17.07} / \underline{0.5179} / 0.4260
            & 16.46 / 0.4110 / 0.4992
            & 16.03 / \textbf{0.5468} / \underline{0.4207}
            & 15.65 / 0.3063 / 0.5376
            & 14.57 / 0.3507 / 0.4954
            & 15.21 / 0.3630 / 0.5423
            & 14.83 / 0.3561 / 0.4838
            & 16.90 / 0.4306 / 0.4350
            & \textbf{17.46} / 0.4436 / \textbf{0.4120}  \\
        gallery chair & 5, 23 & 1, 3, ..., 99
            & 19.78 / 0.6557 / 0.3587
            & \textbf{25.12} / \textbf{0.8192} / \textbf{0.2625}
            & 18.96 / 0.3871 / 0.7146
            & \underline{22.96} / \underline{0.7780} / 0.3349
            & 19.19 / 0.4497 / 0.5694
            & 17.62 / 0.5742 / 0.4504
            & 19.57 / 0.6800 / 0.4585
            & 20.26 / 0.6788 / 0.4059
            & 22.30 / 0.7374 / 0.3264
            & 22.89 / 0.7427 / \underline{0.3060}  \\
        gallery park & 33, 99 & 1, 3, ..., 99
            & 16.29 / 0.5766 / 0.3848
            & 19.19 / \underline{0.6971} / 0.3601
            & 17.87 / 0.3470 / 0.7216
            & \underline{19.67} / \textbf{0.7169} / \underline{0.3562}
            & 18.65 / 0.4561 / 0.5442
            & 17.54 / 0.5788 / 0.4490
            & 14.04 / 0.5440 / 0.5199
            & 17.99 / 0.6136 / 0.4134
            & 19.64 / 0.6568 / 0.3670
            & \textbf{20.13} / 0.6653 / \textbf{0.3409}  \\
        gallery pillar & 21, 39 & 1, 3, ..., 99
            & 19.67 / 0.6461 / 0.3664
            & \underline{21.80} / \textbf{0.7497} / \underline{0.3213}
            & 20.00 / 0.3621 / 0.7285
            & 19.03 / 0.6873 / 0.3694
            & 17.23 / 0.3913 / 0.5875
            & 17.16 / 0.5931 / 0.4376
            & 19.70 / 0.6556 / 0.4402
            & 19.33 / 0.6451 / 0.3821
            & 21.45 / 0.7052 / 0.3477
            & \textbf{22.51} / \underline{0.7284} / \textbf{0.3048}  \\
        garden & 31, 59 & 1, 3, ..., 99
            & 16.97 / 0.5338 / 0.4027
            & 20.38 / \underline{0.6595} / 0.3989
            & 19.22 / 0.3886 / 0.6406
            & \underline{23.09} / \textbf{0.7193} / \underline{0.3373}
            & 18.84 / 0.3943 / 0.5652
            & 15.53 / 0.4690 / 0.4790
            & 16.13 / 0.5098 / 0.5158
            & 20.73 / 0.5723 / 0.4244
            & 22.43 / 0.6335 / 0.3406
            & \textbf{23.45} / 0.6544 / \textbf{0.3052}  \\
        poster & 51, 85 & 1, 3, ..., 99
            & 18.46 / 0.5342 / 0.4219
            & \textbf{20.15} / \textbf{0.6964} / \textbf{0.3263}
            & 15.72 / 0.4675 / 0.5827
            & 17.99 / 0.6179 / 0.4047
            & 16.01 / 0.3879 / 0.5911
            & 14.36 / 0.4897 / 0.4582
            & 13.61 / 0.4881 / 0.5847
            & 17.22 / 0.5780 / 0.4435
            & 18.71 / 0.6517 / 0.3761
            & \underline{19.36} / \underline{0.6674} / \underline{0.3551}  \\
        \midrule
        \midrule
        average & - & - 
            & 17.51 / 0.5309 / 0.3911
            & 19.36 / \underline{0.6307} / 0.3626
            & 16.89 / 0.4113 / 0.5927
            & 18.99 / \textbf{0.6394} / 0.3726
            & 17.06 / 0.3878 / 0.5539
            & 15.68 / 0.4880 / 0.4524
            & 15.85 / 0.5012 / 0.5100
            & 17.88 / 0.5482 / 0.4123
            & \underline{19.70} / 0.6104 / \underline{0.3598}
            & \textbf{20.36} / 0.6288 / \textbf{0.3303}  \\
        \bottomrule
    \end{tabular}
    }
    \caption{Scene-wise quantitative results of 3D reconstruction on \textbf{Ricoh360} dataset.
    }
    \label{tab:ricoh_performance_comparison}
\end{table*}

\begin{table*}[h]
    \centering
    \resizebox{\textwidth}{!}{
    \begin{tabular}{c|c|c|c||c|c|c|c|c|c|c|c|c}
        \toprule[1.0pt]
        Scene & Ref. view & Test view & ODGS & PixelSplat (P) & LatentSplat (P) & MVSplat (P) & PixelSplat (O) & MVSplat (O) & DepthSplat (O) & MVSplat (Y) & OmniSplat & OmniSplat+\textit{opt} \\
        \midrule
        Ballintoy & 0, 35 & 0, 5, ..., 90
            & 19.89 / 0.5674 / 0.4155
            & \textbf{21.78} / 0.7294 / \underline{0.2729}
            & 17.89 / 0.6572 / 0.3928
            & 20.46 / 0.7433 / 0.2817
            & 18.14 / 0.4102 / 0.5756  
            & 18.45 / 0.7034 / 0.3089  
            & 20.34 / 0.7056 / 0.3548
            & 20.12 / 0.7628 / 0.3306
            & 19.07 / \underline{0.7751} / 0.3469  
            & \underline{21.46} / \textbf{0.8076} / \textbf{0.2580}  \\
        BeihaiPark & 0, 55 & 0, 5, ..., 80
            & 21.38 / 0.6417 / 0.2837
            & \textbf{20.37} / \textbf{0.6286} / \textbf{0.3051}
            & \underline{17.64} / 0.5379 / 0.3767
            & 16.86 / 0.4977 / 0.4116
            & 16.89 / 0.4015 / 0.5363  
            & 14.06 / 0.4564 / 0.4534  
            & 16.32 / 0.4822 / 0.4371
            & 16.52 / 0.4948 / 0.4202
            & 17.25 / 0.5324 / 0.3989  
            & 17.62 / \underline{0.5432} / \underline{0.3733}  \\
        Cathedral & 0, 50 & 0, 5, ..., 80
            & 19.54 / 0.5006 / 0.3809
            & \textbf{17.08} / \underline{0.5528} / \textbf{0.3270}
            & 13.50 / 0.3863 / 0.4850
            & 15.29 / 0.4623 / 0.4196
            & 14.32 / 0.2504 / 0.6097  
            & 13.79 / 0.4555 / 0.4124  
            & 14.64 / 0.4180 / 0.4833
            & 15.12 / 0.4402 / 0.4537
            & 15.71 / 0.5288 / 0.4045  
            & \underline{16.24} / \textbf{0.5658} / \underline{0.3672}  \\
        Coast & 0, 45 & 0, 5, ..., 80
            & 21.03 / 0.4943 / 0.4314
            & \textbf{22.53} / 0.7179 / 0.2839
            & 17.95 / 0.5629 / 0.4724
            & \underline{21.74} / \textbf{0.7544} / \underline{0.2828}
            & 17.78 / 0.3252 / 0.5968  
            & 18.02 / 0.6409 / 0.3138  
            & 19.54 / 0.5807 / 0.4244
            & 19.43 / 0.6724 / 0.3382
            & 19.89 / 0.6807 / 0.3580  
            & 21.64 / \underline{0.7368} / \textbf{0.2615}  \\
        Field & 0, 45 & 0, 5, ..., 75
            & 21.10 / 0.5530 / 0.4285
            & \textbf{25.70} / \textbf{0.7509} / \textbf{0.2636}
            & 19.96 / 0.6498 / 0.4201
            & \underline{25.54} / \textbf{0.7584} / \underline{0.2970}
            & 19.61 / 0.3841 / 0.5985  
            & 20.04 / 0.6391 / 0.3745  
            & 21.40 / 0.6279 / 0.4337
            & 23.14 / 0.6644 / 0.3847
            & 22.20 / 0.6903 / 0.4038  
            & 25.15 / 0.7116 / 0.2986  \\
        Nunobiki2 & 0, 50 & 0, 5, ..., 80
            & 19.62 / 0.5883 / 0.3876
            & \textbf{20.03} / \textbf{0.6641} / \textbf{0.3096}
            & 16.47 / 0.5805 / 0.4115
            & 18.15 / 0.6152 / 0.3897
            & 18.03 / 0.4017 / 0.5563  
            & 16.67 / 0.5549 / 0.4363  
            & 19.15 / 0.5970 / 0.4193
            & 18.91 / 0.5965 / 0.3950
            & 18.99 / 0.6267 / 0.3906  
            & \underline{19.68} / \underline{0.6519} / \underline{0.3415}  \\
        SecretGarden1 & 0, 40 & 0, 5, ..., 75
            & 20.91 / 0.6096 / 0.3470
            & 19.57 / 0.6675 / \underline{0.3021}
            & 17.00 / 0.5812 / 0.4043
            & 17.82 / 0.6030 / 0.3741
            & 16.50 / 0.4073 / 0.5424  
            & 15.13 / 0.5545 / 0.4298  
            & 17.44 / 0.5730 / 0.4345
            & 17.85 / 0.6028 / 0.3968
            & \underline{19.71} / \underline{0.6814} / 0.3370  
            & \textbf{20.71} / \textbf{0.7056} / \textbf{0.2926}  \\
        Shrines1 & 0, 45 & 0, 5, ..., 90
            & 18.34 / 0.4264 / 0.3979
            & \underline{16.88} / \textbf{0.4936} / \textbf{0.4174}
            & 13.35 / 0.3260 / 0.5337
            & 15.71 / 0.4179 / 0.5271
            & 14.56 / 0.2643 / 0.5913  
            & 14.35 / 0.3434 / 0.4814  
            & 14.67 / 0.3491 / 0.5763
            & 15.39 / 0.3655 / 0.5163
            & 16.75 / 0.4375 / 0.4634  
            & \underline{17.37} / \underline{0.4712} / \underline{0.4282} \\
        Temple3 & 0, 25 & 0, 5, ..., 70
            & 20.83 / 0.6062 / 0.3167
            & \textbf{17.17} / 0.5869 / \underline{0.3441}
            & 14.21 / 0.4620 / 0.4762
            & 13.63 / 0.4754 / 0.4555
            & 13.02 / 0.3185 / 0.6215  
            & 12.30 / 0.4879 / 0.4152  
            & 14.02 / 0.4896 / 0.5080
            & 14.51 / 0.5221 / 0.4395
            & 16.24 / \underline{0.6164} / 0.3657  
            & \underline{16.67} / \textbf{0.6513} / \textbf{0.3184}  \\
        Wulongting & 0, 50 & 0, 5, ..., 95
            & 19.88 / 0.6722 / 0.3403
            & 18.31 / 0.6938 / \underline{0.2936}
            & 15.01 / 0.6450 / 0.4135
            & 16.04 / 0.6192 / 0.3652
            & 15.56 / 0.3827 / 0.5816  
            & 16.10 / 0.6424 / 0.4236  
            & 16.79 / 0.6466 / 0.4108
            & 16.59 / 0.6461 / 0.3760
            & \underline{19.13} / \underline{0.7417} / 0.3253  
            & \textbf{19.76} / \textbf{0.7633} / \textbf{0.2773}  \\
        \midrule
        \midrule
        average & - & - 
            & 20.25 / 0.5660 / 0.3730
            & \textbf{19.94} / \underline{0.6486} / \textbf{0.3119}
            & 16.30 / 0.5389 / 0.4386
            & 18.12 / 0.5947 / 0.3804
            & 16.44 / 0.3546 / 0.5810  
            & 15.89 / 0.5478 / 0.4049  
            & 17.43 / 0.5470 / 0.4482
            & 17.76 / 0.5768 / 0.4051
            & 18.50 / 0.6311 / 0.3794  
            & \underline{19.63} / \textbf{0.6608} / \underline{0.3217}  \\
        \bottomrule
    \end{tabular}
    }
    \caption{Scene-wise quantitative results of 3D reconstruction on \textbf{OmniPhotos} dataset.
    }
    \label{tab:omniphotos_performance_comparison}
\end{table*}

\begin{table*}[h]
    \centering
    \resizebox{\textwidth}{!}{
    \begin{tabular}{c|c|c|c||c|c|c|c|c|c|c|c|c}
        \toprule[1.0pt]
        Scene & Ref. view & Test view & ODGS & PixelSplat (P) & LatentSplat (P) & MVSplat (P) & PixelSplat (O) & MVSplat (O) & DepthSplat (O) & MVSplat (Y) & OmniSplat & OmniSplat+\textit{opt} \\
        \midrule
        bar & 6, 10 & 7, 8, 9 
            & 17.59 / 0.5680 / 0.3151
            & \underline{14.25} / 0.3583 / \underline{0.4582}
            & 13.76 / \underline{0.3802} / 0.5071
            & 13.42 / 0.3278 / 0.5498
            & 13.16 / 0.2092 / 0.6602  
            & 11.90 / 0.2472 / 0.5911  
            & 12.49 / 0.2652 / 0.6354
            & 13.48 / 0.3211 / 0.5363
            & 13.28 / 0.2862 / 0.5827  
            & \textbf{16.90} / \textbf{0.6174} / \textbf{0.4315}  \\
        base & 21, 25 & 22, 23, 24
            & 19.37 / 0.5900 / 0.3069
            & \textbf{16.04} / \textbf{0.4402} / \textbf{0.4162}
            & 14.03 / 0.4357 / 0.4664
            & \underline{15.79} / \underline{0.4391} / 0.4723
            & 13.48 / 0.1934 / 0.6492  
            & 12.42 / 0.3149 / 0.4942  
            & 14.62 / 0.3606 / 0.5305
            & 14.69 / 0.3652 / 0.5062
            & 15.41 / 0.4231 / 0.4593  
            & 15.27 / 0.3996 / \underline{0.4303}  \\
        cafe & 41, 45 & 42, 43, 44
            & 18.41 / 0.5789 / 0.3565
            & 16.95 / 0.4919 / 0.4498
            & 15.52 / 0.4934 / 0.4787
            & 17.17 / 0.4946 / 0.4764
            & 15.05 / 0.2912 / 0.6352  
            & 13.28 / 0.4129 / 0.4828  
            & 16.64 / 0.4424 / 0.5380
            & 16.77 / 0.4486 / 0.5003
            & \underline{18.73} / \underline{0.5644} / \underline{0.4220}  
            & \textbf{19.05} / \textbf{0.5758} / \textbf{0.3976}  \\
        canteen & 74, 78 & 75, 76, 77
            & 16.42 / 0.4282 / 0.4944
            & 13.37 / 0.4187 / 0.5467
            & 12.94 / \underline{0.4511} / 0.5565
            & 13.39 / 0.4411 / 0.5745
            & 13.44 / 0.2678 / 0.6525  
            & 10.99 / 0.2637 / 0.6211  
            & 13.55 / 0.4009 / 0.6010
            & 14.38 / 0.4320 / 0.5675
            & \textbf{15.39} / \textbf{0.4578} / \underline{0.5393}  
            & \underline{15.16} / 0.4033 / \textbf{0.5318}  \\
        center & 52, 56 & 53, 54, 55
            & 21.07 / 0.6161 / 0.3909
            & 17.17 / 0.5947 / 0.4146
            & 16.70 / 0.6133 / 0.4874
            & 16.91 / 0.5933 / 0.4656
            & 16.07 / 0.3386 / 0.6366  
            & 11.95 / 0.4015 / 0.5175  
            & 17.15 / 0.5602 / 0.5346
            & 18.11 / 0.6064 / 0.4883
            & \underline{20.48} / \textbf{0.6726} / \underline{0.4130}  
            & \textbf{21.08} / \underline{0.6652} / \textbf{0.3642}  \\
        center1 & 36, 40 & 37, 38, 39
            & 19.37 / 0.6232 / 0.4377
            & 16.33 / 0.6594 / 0.4464
            & 16.12 / 0.6832 / 0.4784
            & 16.35 / 0.6534 / 0.4868
            & 15.31 / 0.2966 / 0.6798  
            & 12.10 / 0.3938 / 0.5798  
            & 18.03 / 0.6464 / 0.5146
            & 17.39 / 0.6279 / 0.5318
            & \underline{20.59} / \textbf{0.7127} / \underline{0.4459}  
            & \textbf{20.69} / \underline{0.6956} / \textbf{0.4089}  \\
        corridor & 10, 14 & 11, 12, 13
            & 21.10 / 0.6299 / 0.3130
            & 19.58 / \textbf{0.6934} / \underline{0.3557}
            & 18.15 / 0.6235 / 0.4057
            & 18.27 / \underline{0.6533} / 0.4434
            & 16.19 / 0.2591 / 0.6299  
            & 14.30 / 0.4323 / 0.4587  
            & 18.15 / 0.5251 / 0.5037
            & 18.81 / 0.5449 / 0.4573
            & \underline{20.42} / 0.6104 / 0.3863  
            & \textbf{20.56} / 0.6163 / \textbf{0.3396}  \\
        innovation & 32, 36 & 33, 34, 34
            & 18.29 / 0.5228 / 0.3420
            & 14.73 / \underline{0.3739} / \underline{0.4540}
            & 13.27 / 0.3198 / 0.4987
            & 14.55 / 0.3808 / 0.5127
            & 13.66 / 0.1642 / 0.6266  
            & 13.75 / 0.3093 / 0.4869  
            & 14.05 / 0.2747 / 0.5405
            & 14.73 / 0.3174 / 0.5182
            & \underline{16.61} / \textbf{0.3821} / 0.4555  
            & \textbf{16.82} / 0.3715 / \textbf{0.4378} \\
        lab & 62, 66 & 63, 64, 65
            & 18.13 / 0.5870 / 0.3787
            & 15.23 / 0.4755 / 0.5010
            & 14.89 / 0.5371 / 0.4853
            & 15.40 / 0.5059 / 0.5214
            & 14.76 / 0.2766 / 0.6564  
            & 12.34 / 0.3971 / 0.5288  
            & 16.19 / 0.5221 / 0.5501
            & 16.96 / 0.5625 / 0.4803
            & \underline{18.05} / \textbf{0.6176} / \underline{0.4208}  
            & \textbf{18.23} / \underline{0.6027} / \underline{0.3992}  \\
        library & 5, 9 & 7, 8, 9
            & 19.75 / 0.5127 / 0.4447
            & 18.15 / 0.6126 / \underline{0.4171}
            & 17.76 / 0.6076 / 0.4404
            & 17.73 / \underline{0.6380} / 0.4537
            & 17.06 / 0.3046 / 0.6101  
            & 13.66 / 0.4382 / 0.4566  
            & 18.93 / 0.5583 / 0.4602
            & 20.12 / 0.5751 / 0.4483
            & \underline{20.95} / 0.6119 / 0.4281  
            & \textbf{22.56} / \textbf{0.6567} / \textbf{0.3440}  \\
        office & 80, 84 & 81, 82, 83
            & 16.45 / 0.5361 / 0.4363
            & \underline{17.57} / \underline{0.6789} / \textbf{0.3968}
            & \textbf{18.22} / \textbf{0.7016} / \underline{0.3998}
            & 16.64 / 0.6463 / 0.4661
            & 15.42 / 0.2743 / 0.6368  
            & 11.70 / 0.4372 / 0.5255  
            & 16.15 / 0.5400 / 0.5207
            & 16.51 / 0.5441 / 0.5142
            & 16.76 / 0.5635 / 0.5037  
            & 16.74 / 0.5516 / 0.4740  \\
        \midrule
        \midrule
        average & - & - 
            & 18.72 / 0.5630 / 0.3833 
            & 16.31 / 0.5270 / \underline{0.4415}
            & 15.58 / 0.5315 / 0.4731
            & 15.96 / 0.5249 / 0.4930 
            & 14.87 / 0.2614 / 0.6430  
            & 12.58 / 0.3680 / 0.5221  
            & 16.00 / 0.4633 / 0.5390
            & 16.54 / 0.4859 / 0.5044
            & \underline{17.88} / \underline{0.5366} / 0.4597  
            & \textbf{18.46} / \textbf{0.5596} / \textbf{0.4144}  \\
        \bottomrule
    \end{tabular}
    }
    \caption{Scene-wise quantitative results of 3D reconstruction on \textbf{360Roam} dataset.
    }
    \label{tab:360roam_performance_comparison}
\end{table*}

\begin{table*}[h]
    \centering
    \resizebox{\textwidth}{!}{
    \begin{tabular}{c|c|c|c||c|c|c|c|c|c|c|c|c}
        \toprule[1.0pt]
        Scene & Ref. view & Test view & ODGS & PixelSplat (P) & LatentSplat (P) & MVSplat (P) & PixelSplat (O) & MVSplat (O) & DepthSplat (O) & MVSplat (Y) & OmniSplat & OmniSplat+\textit{opt} \\
        \midrule
        pyebaekRoom & 96, 99 & 97, 98
            & 18.78 / 0.6817 / 0.2925
            & 17.84 / 0.5630 / 0.4072
            & 16.09 / 0.5821 / 0.4428
            & 16.75 / 0.5299 / 0.4445
            & 15.87 / 0.3648 / 0.6134  
            & 14.54 / 0.6011 / 0.3862  
            & 15.97 / 0.5284 / 0.5077
            & 18.12 / 0.5682 / 0.4246 
            & \underline{20.12} / \underline{0.6637} / \underline{0.3274}  
            & \textbf{20.30} / \textbf{0.6786} / \textbf{0.3185}  \\
        room1 & 5, 8 & 6, 7
            & 19.26 / 0.7694 / 0.2868
            & 19.19 / 0.7689 / 0.3240
            & 16.49 / 0.7528 / 0.3613
            & 18.18 / 0.7174 / 0.3724
            & 17.25 / 0.4120 / 0.6108  
            & 12.12 / 0.7739 / 0.3692  
            & 19.38 / 0.7542 / 0.3993
            & 20.81 / 0.7944 / 0.3331
            & \underline{24.05} / \underline{0.8571} / \underline{0.2519}  
            & \textbf{25.73} / \textbf{0.8717} / \textbf{0.2277}  \\
        room2 & 10, 13 & 11, 12
            & 20.71 / 0.7543 / 0.2579
            & 19.93 / 0.7357 / 0.3360
            & 17.68 / 0.7133 / 0.3427
            & 18.27 / 0.7005 / 0.3588
            & 17.76 / 0.3840 / 0.6038  
            & 13.04 / 0.7107 / 0.3559  
            & 18.92 / 0.6761 / 0.4174
            & 21.20 / 0.7367 / 0.3271
            & \underline{21.93} / \underline{0.7894} / \underline{0.2814}  
            & \textbf{23.84} / \textbf{0.8026} / \textbf{0.2595}  \\
        room3 & 90, 93 & 91, 92
            & 21.55 / 0.8191 / 0.2820
            & 22.03 / 0.8157 / 0.3278
            & 19.95 / 0.8115 / 0.3530
            & 21.20 / 0.7924 / 0.3399
            & 18.96 / 0.4393 / 0.6364  
            & 12.96 / 0.7829 / 0.3860  
            & 17.50 / 0.7278 / 0.5128
            & 21.83 / 0.8089 / 0.3601
            & \underline{25.55} / \underline{0.8623} / \underline{0.2872}  
            & \textbf{26.82} / \textbf{0.8650} / \textbf{0.2747}  \\
        room4 & 18, 21 & 19, 20
            & 20.65 / 0.7591 / 0.2843
            & 20.19 / 0.7068 / 0.3651
            & 18.44 / 0.7216 / 0.3971
            & 18.60 / 0.6858 / 0.3808
            & 17.14 / 0.3827 / 0.6304  
            & 13.81 / 0.7421 / 0.3575 
            & 17.64 / 0.6434 / 0.4866
            & 19.85 / 0.7326 / 0.3513
            & \underline{24.14} / \underline{0.8359} / \underline{0.2526}  
            & \textbf{25.81} / \textbf{0.8529} / \textbf{0.2308}  \\
        room5 & 68, 71 & 69, 70
            & 19.97 / 0.7255 / 0.3492
            & 19.76 / 0.7514 / 0.3594
            & 15.72 / 0.7254 / 0.4312
            & 19.64 / 0.7239 / 0.3998
            & 17.34 / 0.3963 / 0.6328  
            & 15.65 / 0.6958 / 0.4013  
            & 19.94 / 0.7758 / 0.3836
            & 20.56 / 0.7467 / 0.4095
            & \underline{24.01} / \textbf{0.8112} / \underline{0.3123}  
            & \textbf{25.19} / \underline{0.8073} / \textbf{0.3000}  \\
        weddingHall & 11, 14 & 12, 13
            & 24.07 / 0.8093 / 0.1716
            & \underline{21.65} / 0.7356 / 0.2989
            & 16.47 / 0.5837 / 0.3977
            & 18.69 / 0.6409 / 0.3689
            & 18.75 / 0.5276 / 0.5298  
            & 13.61 / 0.7346 / \underline{0.2883} 
            & 18.75 / 0.6348 / 0.3918
            & 18.87 / 0.6489 / 0.3554
            & 20.30 / \underline{0.7371} / 0.2891  
            & \textbf{23.35} / \textbf{0.8149} / \textbf{0.2141}  \\
        \midrule
        \midrule
        average & - & - 
            & 20.71 / 0.7598 / 0.2749 
            & 20.08 / 0.7253 / 0.3455
            & 17.27 / 0.6986 / 0.3894
            & 18.76 / 0.6844 / 0.3807
            & 17.58 / 0.4152 / 0.6082  
            & 13.68 / 0.7202 / 0.3635 
            & 18.30 / 0.6772 / 0.4427
            & 20.18 / 0.7195 / 0.3659
            & \underline{22.87} / \underline{0.7938} / \underline{0.2860}  
            & \textbf{24.43} / \textbf{0.8133} / \textbf{0.2608}  \\
        \bottomrule
    \end{tabular}
    }
    \caption{Scene-wise quantitative results of 3D reconstruction on \textbf{OmniScenes} dataset.
    }
    \label{tab:omniscenes_performance_comparison}
\end{table*}

\begin{table*}[h]
    \centering
    \resizebox{\textwidth}{!}{
    \begin{tabular}{c|c|c|c||c|c|c|c|c|c|c|c|c}
        \toprule[1.0pt]
        Scene & Ref. view & Test view & ODGS & PixelSplat (P) & LatentSplat (P) & MVSplat (P) & PixelSplat (O) & MVSplat (O) & DepthSplat (O) & MVSplat (Y) & OmniSplat & OmniSplat+\textit{opt} \\
        \midrule
        seq0 & 25, 28 & 26, 27
            & 17.44 / 0.6157 / 0.3963
            & 20.46 / 0.7155 / 0.2907
            & 19.37 / 0.6965 / 0.3181
            & 20.30 / 0.6915 / 0.2880
            & 17.94 / 0.4881 / 0.5373  
            & 17.56 / 0.6701 / 0.3419  
            & 18.94 / 0.6708 / 0.3379
            & 21.62 / 0.8060 / 0.2396
            & \underline{20.72} / \textbf{0.8171} / \textbf{0.2392}  
            & \textbf{21.45} / \underline{0.7853} / \underline{0.2576}  \\
        seq1 & 164, 167 & 165, 166
            & 24.62 / 0.8860 / 0.1561
            & \underline{18.41} / \textbf{0.6753} / 0.3328
            & 17.86 / 0.6662 / 0.3450
            & \textbf{18.55} / 0.6557 / \underline{0.3281}
            & 17.15 / 0.4186 / 0.5824  
            & 15.55 / 0.5942 / 0.3837  
            & 18.20 / 0.6284 / 0.3758
            & 18.09 / 0.6494 / 0.3367
            & 18.35 / \underline{0.6652} / 0.3308  
            & 18.39 / 0.6649 / \textbf{0.2988}   \\
        seq2 & 72, 75 & 73, 74
            & 31.33 / 0.9537 / 0.0601
            & \underline{20.78} / 0.6551 / 0.3034
            & 19.57 / 0.6527 / 0.3063
            & 19.12 / 0.5537 / 0.3321
            & 20.29 / 0.5304 / 0.5023  
            & 18.45 / \underline{0.7113} / \underline{0.2340}  
            & 19.85 / 0.5951 / 0.3341
            & 20.57 / 0.6976 / 0.2373
            & 20.42 / 0.7054 / 0.2465  
            & \textbf{22.61} / \textbf{0.7716} / \textbf{0.1930} \\
        seq3 & 25, 28 & 26, 27
            & 20.62 / 0.7533 / 0.2109
            & \textbf{18.50} / 0.5890 / 0.3410
            & 17.20 / 0.5877 / 0.3579
            & 17.47 / 0.5087 / 0.3965
            & 17.12 / 0.4645 / 0.5162  
            & 15.55 / 0.6107 / 0.3087  
            & 16.47 / 0.4991 / 0.4380
            & 17.53 / 0.6214 / 0.2989
            & 18.15 / \underline{0.6327} / \underline{0.2962}  
            & \underline{18.47} / \textbf{0.6814} / \textbf{0.2470} \\
        seq4 & 41, 44 & 42, 43
            & 26.72 / 0.8253 / 0.1464
            & 20.24 / 0.6596 / 0.2824
            & 19.66 / 0.6375 / 0.3007
            & 19.15 / 0.5719 / 0.3054
            & 18.74 / 0.4495 / 0.5132  
            & 19.45 / 0.6879 / 0.2465  
            & 18.71 / 0.5671 / 0.3559
            & 19.63 / 0.6475 / 0.2611
            & \underline{21.08} / \underline{0.6928} / \underline{0.2359}  
            & \textbf{23.09} / \textbf{0.7498} / \textbf{0.1882} \\
        seq5 & 14, 17 & 15, 16
            & 18.40 / 0.7364 / 0.2815
            & 18.55 / 0.7049 / 0.3106
            & 17.00 / 0.6772 / 0.3342
            & \textbf{19.08} / 0.6894 / 0.3214
            & 17.38 / 0.5207 / 0.5565  
            & 16.61 / 0.6933 / 0.3467  
            & 17.56 / 0.6765 / 0.3734
            & 17.81 / 0.6871 / 0.3175
            & 18.73 / \underline{0.7543} / \underline{0.2873}  
            & \underline{18.98} / \textbf{0.7746} / \textbf{0.2643} \\
        seq6 & 12, 15 & 13, 14
            & 22.49 / 0.7826 / 0.2296
            & 18.23 / 0.5809 / 0.3203
            & 18.13 / 0.5735 / 0.3494
            & 19.08 / 0.5781 / 0.2999
            & 17.54 / 0.3615 / 0.5585  
            & 20.02 / 0.7005 / 0.2552  
            & 18.98 / 0.5407 / 0.3889
            & 20.36 / 0.6531 / 0.2514
            & \underline{20.76} / \underline{0.7103} / \underline{0.2274}  
            & \textbf{23.76}/ \textbf{0.7680} / \textbf{0.1852} \\
        seq7 & 221, 224 & 222, 223
            & 22.87 / 0.8924 / 0.1087
            & 19.94 / 0.7249 / 0.2713
            & 17.00 / 0.6574 / 0.3258
            & 20.81 / 0.7431 / 0.2083
            & 18.71 / 0.5261 / 0.5264  
            & 20.97 / \underline{0.8755} / \underline{0.1389}  
            & 18.88 / 0.6916 / 0.2876 
            & \underline{22.60} / 0.8372 / 0.1637
            & 21.39 / 0.8611 / 0.1705  
            & \textbf{25.02} / \textbf{0.9296} / \textbf{0.0863} \\
        seq8 & 42, 45 & 43, 44
            & 20.15 / 0.5901 / 0.3232
            & 18.42 / 0.5803 / 0.3191
            & 18.09 / 0.5639 / 0.3778
            & 19.47 / 0.6065 / 0.3184
            & 16.77 / 0.4288 / 0.5575  
            & 15.85 / 0.5528 / 0.3442  
            & 20.45 / 0.6112 / 0.3498
            & 20.40 / 0.6040 / 0.2937
            & \underline{20.52} / \textbf{0.6406} / \underline{0.2849}  
            & \textbf{20.80} / \underline{0.6372} / \textbf{0.2581} \\
        seq9 & 134, 137 & 135, 136
            & 21.95 / 0.7509 / 0.3089
            & 19.77 / 0.7423 / \underline{0.3079}
            & 19.68 / \underline{0.7441} / 0.3294
            & 18.92 / 0.7236 / 0.3199
            & 16.27 / 0.4464 / 0.5603  
            & 14.51 / 0.5982 / 0.3916  
            & 18.92 / 0.6890 / 0.3991
            & \underline{19.90} / 0.7184 / 0.3305
            & \textbf{20.57} / \textbf{0.7458} / \textbf{0.2993}  
            & 19.13 / 0.7173 / 0.3262 \\
        \midrule
        \midrule
        average & - & - 
            & 22.66 / 0.7786 / 0.2222 
            & 19.33 / 0.6628 / 0.3079
            & 18.36 / 0.6457 / 0.3345
            & 19.19 / 0.6322 / 0.3118
            & 17.79 / 0.4635 / 0.5411  
            & 17.45 / 0.6695 / 0.2991 
            & 18.69 / 0.6170/ 0.3641
            & 19.85 / 0.6922 / 0.2730
            & \underline{20.07} / \underline{0.7225} / \underline{0.2618}  
            & \textbf{21.17} / \textbf{0.7480} / \textbf{0.2305} \\
        \bottomrule
    \end{tabular}
    }
    \caption{Scene-wise quantitative results of 3D reconstruction on \textbf{360VO} dataset.
    }
    \label{tab:360vo_performance_comparison}
\end{table*}


\end{document}